\documentclass{article}

\usepackage[preprint]{neurips_2026}
\usepackage[table,xcdraw]{xcolor}
\usepackage{multirow}

\usepackage{natbib}
\setcitestyle{numbers,square}
\usepackage{amssymb}
\usepackage{xspace}
\usepackage[utf8]{inputenc} 
\usepackage[T1]{fontenc}    
\usepackage[hidelinks]{hyperref}
\usepackage{amsthm}
\usepackage{url}            
\usepackage{booktabs}       
\usepackage{amsfonts}       
\usepackage{nicefrac}       
\usepackage{microtype}      
\usepackage{xcolor}         
\usepackage{graphicx}      
\usepackage{wrapfig}
\usepackage{amsmath} 
\usepackage{booktabs}  
\usepackage{enumitem}
\usepackage{makecell}
\definecolor{headergray}{RGB}{242, 244, 247}
\definecolor{subheadergray}{RGB}{248, 249, 251}
\definecolor{modelgray}{RGB}{235, 238, 242}
\definecolor{ourslight}{RGB}{239, 246, 255}
\definecolor{stdblue}{RGB}{54, 95, 145}
\definecolor{bestblue}{RGB}{20, 70, 120}
\definecolor{retainlight}{RGB}{240, 248, 240}
\newcommand{\scorestd}[2]{#1\,{\scriptsize\textcolor{stdblue}{$\pm$#2}}}
\newcommand{\bestscorestd}[2]{\textbf{#1}\,{\scriptsize\textcolor{stdblue}{$\pm$#2}}}
\newcommand{\eg}{\textit{e.g.}\@\xspace}
\newcommand{\ie}{\textit{i.e.}\@\xspace}
\theoremstyle{definition}           
\usepackage{multirow}
\usepackage{physics} 
\usepackage{tcolorbox}
\usepackage{hyperref}
\usepackage{microtype}
\usepackage{graphicx}
\usepackage{booktabs} 
\usepackage{algorithm}
\usepackage{algorithmic}
\usepackage{microtype}
\usepackage{booktabs} 
\usepackage{todonotes}
\usepackage{multirow} 
\usepackage{amsmath}
\usepackage{mathtools}
\usepackage{amsthm}
\newcommand{\fit}{\texttt{FIT}\xspace}
\def\eg{\textit{e.g.}\xspace}
\def\ie{\textit{i.e.}\xspace}

\usepackage{array}

\makeatletter
\newcommand{\algorithmicpp}{\textbf{PP:}}
\newcommand{\PP}{\item[\algorithmicpp]}
\makeatother
\usepackage{subcaption}
\usepackage{multirow} 
\usepackage{pifont}
\usepackage{amssymb} 
\usepackage{xspace} 
\usepackage{booktabs}  
\usepackage[capitalize,noabbrev]{cleveref}
\usepackage{fontawesome5}
\theoremstyle{plain}

\theoremstyle{definition}

\theoremstyle{remark}

\usepackage{fontawesome5}
\title{\Large \fit to Forget: Robust Continual Unlearning 
\\ for Large Language Models}

\author{%
Xiaoyu Xu$^{\dagger}$,\quad
Minxin Du$^{\dagger}$\thanks{Corresponding authors: Haibo Hu (\texttt{haibo.hu@polyu.edu.hk}) and Minxin Du (\texttt{minxin.du@polyu.edu.hk}).},\quad
Kun Fang$^{\dagger}$,\quad
Yaxin Xiao$^{\dagger}$,\quad
Zhicong Huang$^{\diamond}$\\[2pt]
\textbf{Cheng Hong}$^{\diamond}$,\quad
\textbf{Qingqing Ye}$^{\dagger}$,\quad
\textbf{Haibo Hu}$^{\dagger}$\footnotemark[1]\\
\\
$^{\dagger}$The Hong Kong Polytechnic University\\
$^{\diamond}$Ant Group\\
\\
\texttt{xiaoyu0910.xu@connect.polyu.hk}\\
\faGithub\ \href{https://github.com/XiaoyuXU1/FIT}{\texttt{Code}}
\quad
\faGlobe\ \href{https://xiaoyuxu1.github.io/FIT_PCH/}{\texttt{Website}}
}

\begin{document}

\maketitle
\begin{abstract}
While large language models (LLMs) exhibit remarkable capabilities, they increasingly face demands to unlearn memorized privacy-sensitive, copyrighted, or harmful content. 
Existing unlearning methods primarily focus on \emph{single-shot} scenarios, whereas real-world deletion requests arrive \emph{continually}. 
Na\"ively applying these methods to sequential requests leads to severe utility degradation and catastrophic forgetting. 
To address this, we propose \fit, a robust continual unlearning framework to process high-volume sequential deletion streams while resisting both catastrophic forgetting and post-unlearning recovery. 
\fit stabilizes sequential updates through three synergistic mechanisms: redundancy \underline{F}iltering, \underline{I}mportance-aware adaptive algorithm selection, and \underline{T}argeted layer attribution. 
Furthermore, to facilitate rigorous evaluation, we introduce \textbf{PCH}, a unified benchmark encompassing \textbf{P}ersonal, \textbf{C}opyrighted, and \textbf{H}armful content, alongside two symmetric metrics, Forget Degree (F.D.) and Retain Utility (R.U.), to systematically quantify forgetting-utility trade-offs. Extensive experiments across five LLMs (up to 14B parameters) demonstrate that \fit consistently achieves state-of-the-art unlearning efficacy and utility preservation. 
Notably, even after hundreds of sequential requests, \fit preserves strong downstream (\eg, GSM8K, MMLU) performance and exhibits superior resilience against relearning and quantization recovery attacks.


\end{abstract}

\section{Introduction}{\label{intro}}
Large language models (LLMs) exhibit remarkable versatility but pose significant ethical and legal risks due to their tendency to memorize sensitive, harmful, or copyrighted training data~\cite{emnlp/KaramolegkouLZS23}.
To comply with regulations, such as the GDPR~\cite{mantelero2013} (``Right to be Forgotten'') and CCPA~\cite{harding2019}, \emph{machine unlearning} has emerged as a critical mechanism for erasing specific data influences~\cite{sp/BourtouleCCJTZL21,sp/CaoY15}.

Unlearning can be either exact or approximate. 
\emph{Exact unlearning} requires the unlearned model to match the distribution of a model retrained from scratch on the retain set. 
SISA achieves this by partitioning data into disjoint shards and retraining only the affected shard after deletion~\cite{sp/BourtouleCCJTZL21}, but this is prohibitively expensive for LLMs. 
Recent work therefore focuses on \emph{approximate unlearning}, which only requires statistical or behavioral similarity. 
Representative methods include gradient ascent (GA), random labeling (RLabel), and negative preference optimization (NPO), which can remove targets but often harm utility~\cite{corr/zhang24}. This motivates utility-preserving variants such as GA+GD~\cite{acl/YaoCDNWCY24}.

\begin{wrapfigure}{r}{0.5\textwidth}
    \centering
    \vspace{-10pt}
    \includegraphics[width=0.5\textwidth]{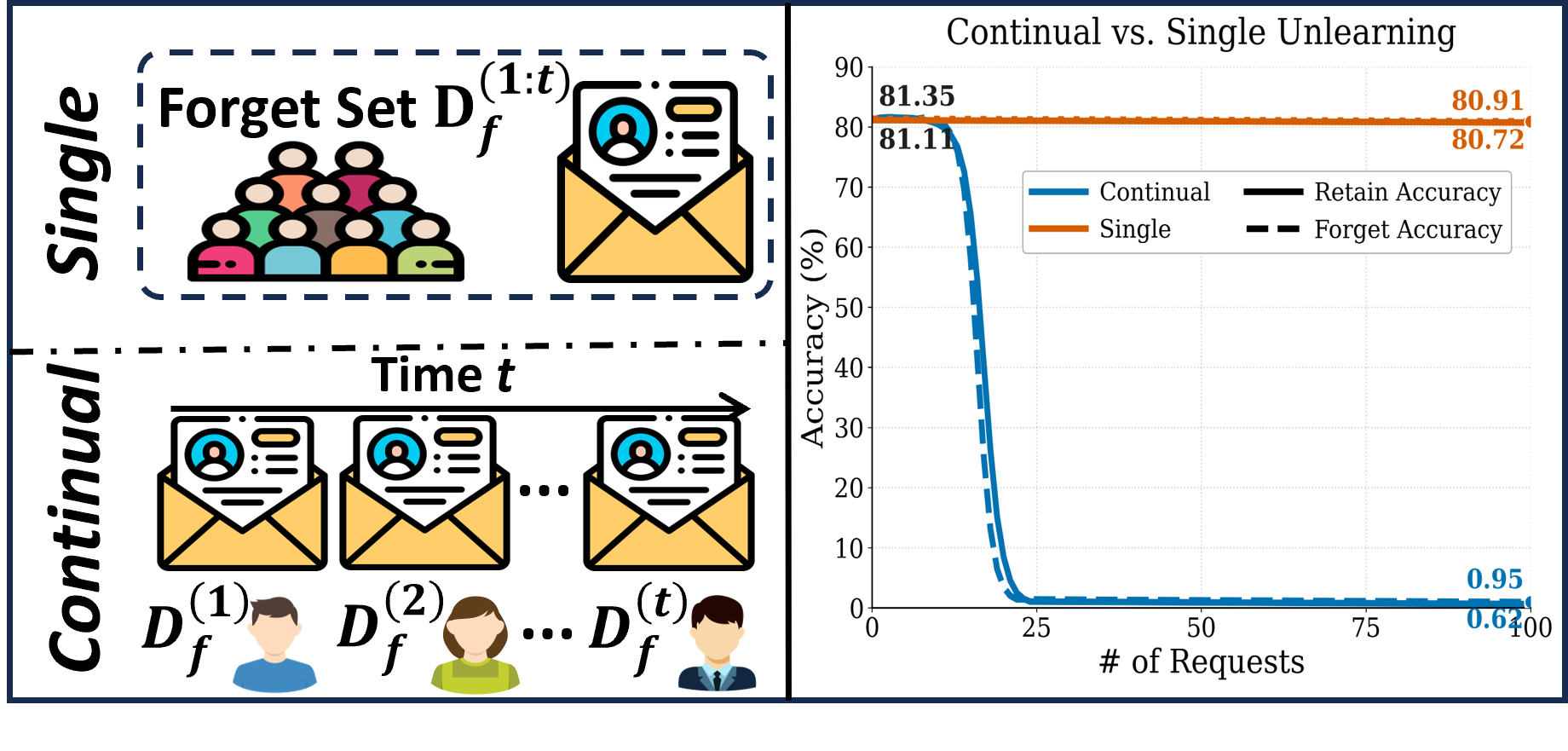}
    \caption{
    \textbf{Left}: Schematics of \emph{single-shot} vs. \emph{continual} unlearning;
    \textbf{Right}: Retain and forget accuracy on Llama-3-8B using GA for single unlearning and $100$ sequential request(s)
    }
    \label{fig:100users_continual}
    \vspace{-10pt}
\end{wrapfigure}

However, they predominantly focus on a \emph{single-shot} setting, where the entire forget set is removed at once. 
In practice, \emph{continual unlearning} arises, where requests arrive sequentially, repeatedly, and in intertwined forms throughout an LLM's life cycle~\cite{natmi/LiuYJCBHYLXLVBKL25}. 
Directly extending them by treating each request independently often results in severe utility degradation or even \emph{catastrophic forgetting}~\cite{iclr/shi24, corr/Barez25, natmi/LiuYJCBHYLXLVBKL25}. 
As shown in Figure~\ref{fig:100users_continual}, while single-shot removal has minimal impact, continual unlearning causes a rapid and cumulative decline in both forget and retain accuracy after only 25 sequential requests. Similar failure modes have also been reported for image models~\cite{corr/Thakral25, icml/lee2025an, CVPR/zhao2024}.

Continual unlearning for LLMs remains underexplored.
Orthogonal unlearning (\(O^3\))~\cite{iclr/gao25} improves efficiency by combining LoRA adapters with an out-of-distribution detector. 
Yet, detector errors can compromise forgetting, while the LoRA parameterization may increase the risk of reactivating erased knowledge~\cite{iclr/0001FWS25}. 
Other approaches, such as ALKN~\cite{icml/wuerkaixi2025adaptive}, mitigate parameter drift through adaptive task vectors but require costly gradient inspection. 
Overall, these methods still struggle with catastrophic forgetting under sustained request streams. 
In addition, unlearned models may remain susceptible to post-unlearning recovery induced by small parameter updates like relearning through fine-tuning~\cite{acl/LoBC24,icml/xu2025} and quantization-based attacks~\cite{iclr/ZhangWLWTL00W25}.

We therefore ask whether there exists a continual unlearning framework, built on standard single-shot methods, that simultaneously satisfies four desiderata: 
\emph{i) the simplicity of single-shot objectives, 
ii)  the efficiency and effectiveness of specialized approaches such as ALKN,
iii) resilience to catastrophic forgetting under long-horizon sequential requests, 
and (iv) robustness to post-unlearning recovery.}

\subsection{Technical Overview}
\begin{figure*}[!t]
    \centering
    \includegraphics[width=0.85\linewidth]{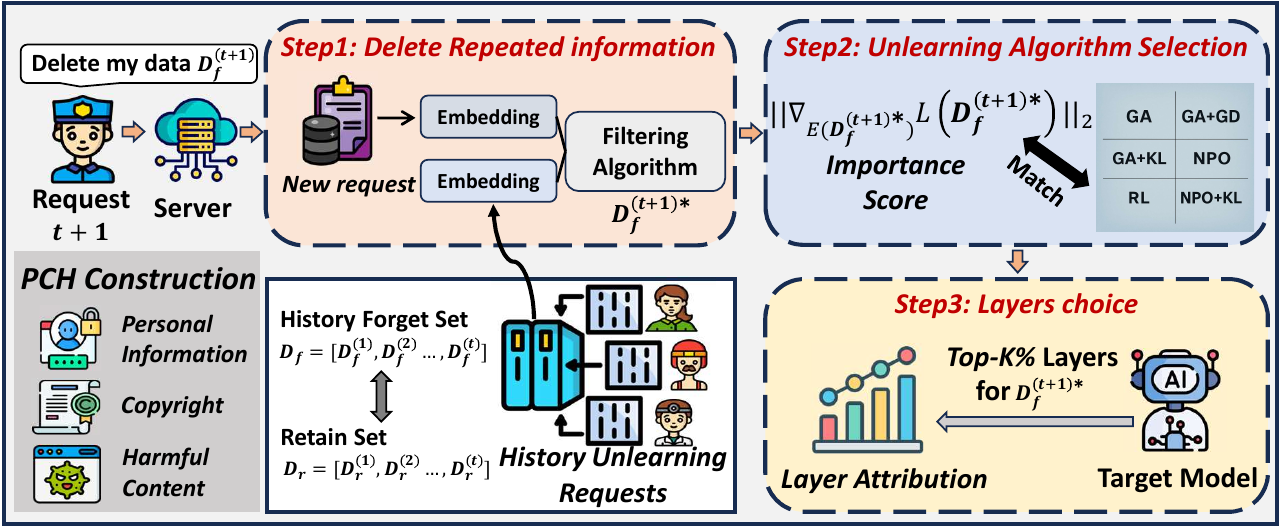}
    \caption{Overview of \fit:
    Incoming unlearning requests are first de-duplicated via embedding-based redundancy filtering (Section~\ref{subsec:redundancy_filtering}).
    For each filtered request, an importance score then guides adaptive selection of the unlearning method (Section~\ref{subsec:algorithm_selection}), and targeted layer attribution restricts updates to the top‑$K\%$ influential layers (Section~\ref{subsec:layer_attribution}), mitigating compounded knowledge loss and parameter drift.}
    \label{fig:overview}
\end{figure*}

We attribute catastrophic forgetting in continual unlearning to three primary drivers: 
i) cumulative redundancy from semantically overlapping requests~\cite{icml/wuerkaixi2025adaptive}, 
ii) step-wise optimization instability~\cite{corr/Barez25}, and iii) long-term parameter drift~\cite{acl/Bae26}. 
Crucially, managing parameter drift involves a delicate trade-off: excessive drift leads to model collapse, whereas insufficient drift leaves the model vulnerable to recovery attacks. 
Guided by these insights, we introduce \fit to overcome these challenges (Figure~\ref{fig:overview}).

First, to mitigate the utility loss of repeatedly processing overlapping requests, \fit employs a two-stage \underline{F}iltering mechanism, combining embedding-based similarity checks with a loss-difference test, to safely prune redundant context while preserving sensitive target information.
Second, to address update instability, we introduce an \underline{I}mportance-aware adaptive mechanism that scores the influence of each request, subsequently routing it to an appropriately aggressive or conservative unlearning algorithm. 
As supported by our theoretical analysis, this stabilizes gradient directions and minimizes collapse risk. 
Finally, to curb excessive parameter drift, we utilize \underline{T}argeted layer attribution based on Shapley-style relevance estimation~\cite{ijcai/RozemberczkiWBY22}. 
By restricting updates to the top-$K\%$ most relevant layers per request, this ratio-based design scales naturally with model size, ensuring request-dependent flexibility while tightly bounding computational overhead and structural degradation.

\subsection{A New Benchmark}
Existing unlearning datasets focus primarily on single-shot settings~\cite{colm/Maini24,icml/LiPGYBGLDGMHLJL24,iclr/shi24,nips/JinCWHYL00024,acl/YaoCDNWCY24}, and each covers only one deletion category (Table~\ref{tab:dataset-overview}). TOFU~\cite{colm/Maini24} evaluates fictitious-author removal for privacy, WMDP~\cite{icml/LiPGYBGLDGMHLJL24} targets hazardous knowledge, and MUSE~\cite{iclr/shi24} focuses on copyright deletion using curated news and book text. Such category-specific designs may encourage unlearning methods to overfit to particular deletion types while missing broader real-world variation, motivating a unified benchmark for continual unlearning. Evaluation is further limited by inconsistent metrics: MUSE uses disparate criteria that obscure overall trade-offs, while TOFU relies on paraphrased answers and mismatched protocols across forget and retain sets, leading to inconsistent results.

To address the lack of a dedicated \emph{continual} unlearning benchmark, we introduce \textbf{PCH}, which unifies \textbf{P}ersonal information, \textbf{C}opyright, and \textbf{H}armful content. All instances are synthetically generated by GPT-4o using structured prompts (without real data or violating OpenAI's usage policies) to reduce overlap with common pre-training corpora, and are manually verified for category consistency and basic distributional properties. These steps enable the construction of faithful retain baselines. 

We further propose two symmetric metrics: Forget Degree (F.D.) and Retain Utility (R.U.). Computed as the geometric mean of three underlying measures, Probability, ROUGE-L, and token-level Accuracy, on the forget and retain sets, they provide a scale-invariant, interpretable assessment of the trade-off between forgetting and retention without allowing any single factor to dominate.

\textbf{Our main contributions are summarized below.}

\noindent I) We propose \fit, a robust and practical continual unlearning framework to defy catastrophic forgetting in LLMs. 
It integrates three strategic mechanisms: embedding-based redundancy \underline{F}iltering to prevent gradient accumulation, \underline{I}mportance-aware adaptive algorithm selection to stabilize sequential updates, and \underline{T}argeted layer attribution to minimize parameter drift.

\noindent II) We introduce \textbf{PCH}, a unified continual unlearning benchmark encompassing \textbf{P}ersonal information, \textbf{C}opyright, and \textbf{H}armful content. 
To overcome the limitations of disparate evaluation criteria, we propose two symmetric metrics, Forget Degree (F.D.) and Retain Utility (R.U.), which provide a \emph{scale-invariant, interpretable} assessment of trade-offs between forgetting efficacy and model utility.

\noindent III) We conduct extensive experiments on five LLMs. 
To our knowledge, we are the \emph{first} to evaluate continual LLM unlearning at a scale of up to $300$ sequential requests, far exceeding the typical ${<}10$ requests~\cite{iclr/gao25,icml/wuerkaixi2025adaptive}, and on models up to $14$B parameters, whereas most prior arts focus on $7$ or $8$B baselines.
Results show that \fit achieves a better forgetting-utility trade-off (\eg, outperforming ALKN and \(O^3\) by up to +0.09 F.D. and +0.08 R.U. on Llama-3-8B), while preserving downstream performance and exhibiting stronger resilience against relearning and quantization attacks.


\section{Problem Formulation and Preliminaries}{\label{pre}}
\subsection{Problem Formulation}\label{problem}
Let $\mathsf{D}$ denote the full (pre-)training corpus and $\mathcal{A}$ the algorithm producing the model $\mathcal{M} = \mathcal{A}(\mathsf{D})$.
Given a \emph{forget set} $\mathsf{D}_f \subset \mathsf{D}$, an unlearning operator $\mathcal{U}$ produces an updated model $\mathcal{M}_f = \mathcal{U}(\mathcal{M}, \mathsf{D}_f)$.
Its complement is the \emph{retain set} $\mathsf{D}_r = \mathsf{D} \setminus \mathsf{D}_f$.
Unlearning generally falls into two categories: \emph{exact} and \emph{approximate}~\cite{sp/BourtouleCCJTZL21}.
Exact unlearning requires $\mathcal{M}_f$ to be distributionally identical to a retrained model $\mathcal{M}_r = \mathcal{A}(\mathsf{D}_r)$, ensuring all statistical traces of $\mathsf{D}_f$ are removed.
However, exact methods (\eg, SISA~\cite{sp/BourtouleCCJTZL21}) are prohibitively expensive for modern LLMs.
We therefore focus on approximate unlearning, which relaxes strict equivalence in favor of behavioral similarity~\cite{acl/YaoCDNWCY24,colm/Maini24,iclr/shi24}.

While most existing work focuses on \emph{single-shot} unlearning~\cite{acl/YaoCDNWCY24,icml/PawelczykNL24,icml/LiPGYBGLDGMHLJL24,usenix/long25,usenix/Song25}, where $\mathcal{U}$ is invoked once for a single forget set $\mathsf{D}_f$.
Instead, we address the more realistic--yet scarcely explored--\emph{continual unlearning} scenario.
Here, unlearning requests $\mathcal{D}_f^{(1)}, \ldots, \mathcal{D}_f^{(t)}$ arrive sequentially, reflecting real-world online deletion demands.
At round $t$, the cumulative forget set is $\mathsf{D}_f^{(1:t)} = \bigcup_{i=1}^t \mathcal{D}_f^{(i)}$, with the corresponding retain set $\mathsf{D}_r^{(1:t)} = \mathsf{D} \setminus \mathsf{D}_f^{(1:t)}$.
The model updates iteratively: $\mathcal{M}_f^{(i)} = \mathcal{U}(\mathcal{M}_f^{(i-1)}, \mathcal{D}_f^{(i)})$.
Na\"ive sequential application of $\mathcal{U}$ often degrades utility rapidly, leading to catastrophic forgetting~\cite{iclr/shi24}.

Our threat model (Appendix~\ref{threat}) considers this instability alongside several practical adversarial risks, including \emph{``malicious'' unlearning} with a large volume of requests (cf. denial of service) to induce collapse~\cite{corr/Barez25}, \emph{relearning attacks} that attempt to recover erased knowledge after unlearning~\cite{acl/LoBC24}, and \emph{quantization attacks} that may restore residual memorization through low-bit compression~\cite{iclr/ZhangWLWTL00W25}.

\textbf{Practical evaluation.}
Ideally, the unlearning quality at round $i$ is measured against a ``gold-standard'' retrained model 
$\mathcal{M}_r^{(i)} = \mathcal{A}(\mathsf{D}_r^{(1:i)})$.
Since full retraining on the unavailable corpus $\mathsf{D}$ is infeasible, we adopt the synthetic proxy approach from~\cite{colm/Maini24}; see Appendix~\ref{Experimental configuration details} for details.
We synthesize disjoint datasets ``$\mathsf{D}_f$'' and ``$\mathsf{D}_r$.''
We then:
i) fine-tune $\mathcal{M}$ on the union ``$\mathsf{D}_f \cup \mathsf{D}_r$'' to embed the knowledge, 
and ii) fine-tune a separate copy of $\mathcal{M}$ solely on ``$\mathsf{D}_r$'' to serve as the \emph{retain model}.
This surrogate provides a rigorous baseline for evaluating both deletion fidelity and utility preservation.



\subsection{Common (Single-shot) Unlearning Methods} 
\label{related work}
We use three classes of single-shot methods as primitives.
\textbf{(i) GA Family:} 
It maximizes loss on $\mathsf{D}_f$ while preserving performance on $\mathsf{D}_r$:
$
  \mathcal{L}
  = \mathcal{L}_{\text{GA}}\bigl(\mathsf{D}_f\bigr)
  + \lambda\,\mathcal{L}_{\text{retain}}\bigl(\mathsf{D}_r\bigr),
$
where $\lambda \geq 0$. 
Variants include pure GA ($\lambda = 0$), GA+GD (with cross-entropy on $\mathsf{D}_r$), and GA+KL (with KL divergence to a reference model)~\cite{acl/YaoCDNWCY24}.
\textbf{(ii) NPO Family:} 
It prevents over-forgetting by penalizing the model's alignment with $\mathsf{D}_f$ rather than maximizing loss~\cite{corr/zhang24}:
$
  \mathcal{L}
  = \mathcal{L}_{\text{NPO}}\bigl(\mathsf{D}_f\bigr)
  + \lambda\,\mathcal{L}_{\text{retain}}\bigl(\mathsf{D}_r\bigr).
$
The NPO+KL variant applies KL divergence on~$\mathsf{D}_r$.
\textbf{(iii) RLabel:} 
It enforces uniform predictions by training on random labels for $\mathsf{D}_f$~\cite{acl/YaoCDNWCY24}:
$
  \mathcal{L}
  = \mathcal{L}_{\text{RLabel}}\bigl(\mathsf{D}_f\bigr).
$ Each method presents a trade-off: 
aggressive strategies like GA ensure forgetting but risk severe utility degradation (or over-forgetting)~\cite{acl/YaoCDNWCY24}. 
Conversely, NPO or NPO+KL better preserve utility but may be less effective at erasing knowledge~\cite{corr/zhang24}.

\section{Our Approach:~\fit}{\label{method}}


Continual LLM unlearning risks catastrophic forgetting as deletion requests accumulate. 
We identify three primary drivers of this collapse: 
i) gradient compounding from redundant requests~\cite{icml/wuerkaixi2025adaptive}, 
ii) optimization instability across sequential steps~\cite{corr/Barez25}, and 
iii) parameter drift from indiscriminate updates~\cite{acl/Bae26}. 
To address these, we propose \fit, a robust continual unlearning framework to mitigate forgetting and post-unlearning recovery via three synergistic modules (Figure~\ref{fig:overview}):

i) Two-Stage \underline{F}iltering: Prevents compounded knowledge loss by pruning redundant requests prior to optimization. 
It combines chunk-level embedding similarity with a loss-difference test to discard repetitive contexts while preserving structurally similar but semantically unique sensitive data.

ii) \underline{I}mportance-Guided Algorithm Selection: Stabilizes sequential updates by dynamically routing requests to appropriately aggressive or conservative unlearning algorithms. It achieves this by using the $L_2$-norm of the embedding gradients as a lightweight proxy for request memorization~\cite{iclr/AnconaCO018}.

iii)  \underline{T}argeted Layer Attribution: Curbs parameter drift by updating only the most relevant subnetworks. 
Using leave-one-out loss deviations to approximate Shapley values~\cite{ijcai/RozemberczkiWBY22}, it dynamically identifies the top-$K\%$ most influential layers per request, restricting updates to their MLP and attention modules.

\subsection{Redundancy Filtering}
\label{subsec:redundancy_filtering}

Unlearning requests from diverse sources often contain semantically redundant text. 
When these overlapping requests are sequentially removed, their aligned gradients accumulate along shared dimensions. 
This leads to the systematic suppression of common tokens rather than the targeted erasure of specific knowledge~\cite{icml/wuerkaixi2025adaptive}. 
As modeled in Appendix~\ref{filtering_proof} and illustrated in Figure~\ref{fig:token_prob_decay}, repeated updates precipitate a collapse of shared-token probabilities toward zero. Once these tokens collapse, the
model loses the semantic distinctions they support, resulting in catastrophic forgetting. Ideally, unlearning mechanisms should prevent this collapse by stabilizing token probabilities at moderate levels, thereby reducing reliance on targeted information while preserving the semantic capacity.

A standard countermeasure is to filter incoming requests based on their embedding similarity to historical data. 
However, semantic similarity does not strictly equate to redundancy. 
Texts may exhibit high lexical overlap while conveying distinct, sensitive information (\eg, ``My name is Alice'' vs.\ ``... Bob'').
Consequently, standard filtering faces a strict trade-off: aggressive pruning risks ignoring legitimately forgettable content (with information leakage), while insufficient pruning exacerbates gradient overlap (hence catastrophic forgetting).
\begin{wrapfigure}{r}{0.51\textwidth}
    \centering
    \vspace{-10pt}
    \includegraphics[width=0.41\textwidth]{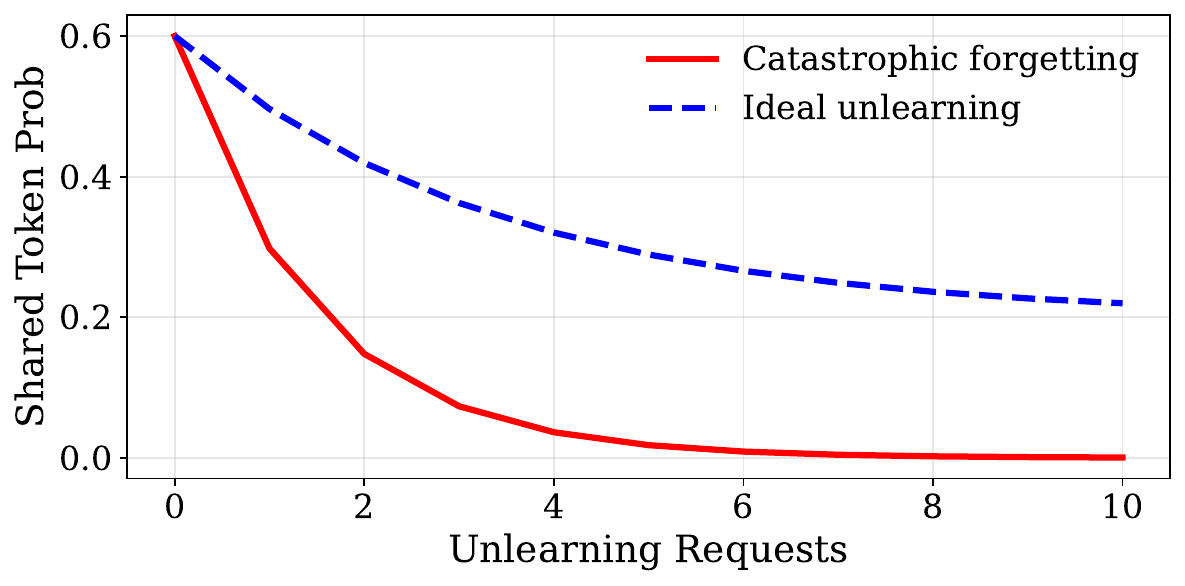}
    \caption{Estimated decay of shared-token probabilities under continual unlearning: Redundant gradients drive probabilities toward zero, inducing catastrophic collapse, whereas effective unlearning maintains them at a stable, moderate level.}
    \label{fig:token_prob_decay}
    \vspace{-35pt}
\end{wrapfigure}
To navigate this, we propose a two-stage filtering protocol that balances redundancy reduction with precise knowledge preservation. 
Given a historical forget set $\mathsf{D}_f^{(1:t)}$ and a new request $\mathcal{D}_f^{(t+1)}$, we construct a filtered request $\mathcal{D}_f^{(t+1)*}$ through the following steps:

Firstly, we partition each sample in $\mathcal{D}^{(t+1)}_f$ into fixed-size chunks, where a chunk refers to an intra-sample segment. 
For each chunk $x$, we compute its SimCSE embedding $\mathsf{e}(x)$~\cite{emnlp/GaoYC21} and determine its maximum cosine similarity $s^*$ against $\mathsf{D}_f^{(1:t)}$.
If $s^*$ falls below a threshold $\tau$, the chunk is designated as non-redundant and marked for unlearning. Secondly, for chunks where $s^* \geq \tau$, we apply a secondary \emph{loss-difference test} to prevent the erroneous discarding of structurally similar but semantically unique data (\eg, distinct personally identifiable information). 
We compute:
\begin{equation}
\Delta L = \left| L_{\text{with}} - L_{\text{without}} \right|,
\end{equation}
where $L_{\text{with}}$ and $L_{\text{without}}$ are the cross-entropy losses evaluated on the full $\mathcal{D}_f^{(t+1)}$ and the ablated set $\mathcal{D}_f^{(t+1)}\setminus{x}$. A significant deviation ($\Delta L > \epsilon$) indicates that $x$ contributes non-trivial information to the model’s predictions, aligned with our analysis in Appendix~\ref{filtering_proof}. Such chunks are retained for unlearning despite their high embedding similarity. See Algorithm~\ref{alg:embedding-filtering} for pseudocode (Appendix~\ref{filtering_proof}).


Empirical validation using GPT-4o sensitivity scoring~\cite{corr/weiGu24} and term-distribution visualizations (Appendix~\ref{filtering}) confirms that this two-stage mechanism safely eliminates redundant context while reliably preserving sensitive target tokens, including names, harmful terms, and copyrighted expressions.

\subsection{Importance-Guided Algorithm Selection}
\label{subsec:algorithm_selection}
\begin{wrapfigure}{r}{0.54\textwidth}
    \centering
    \vspace{-12pt}
    \includegraphics[width=0.53\textwidth]{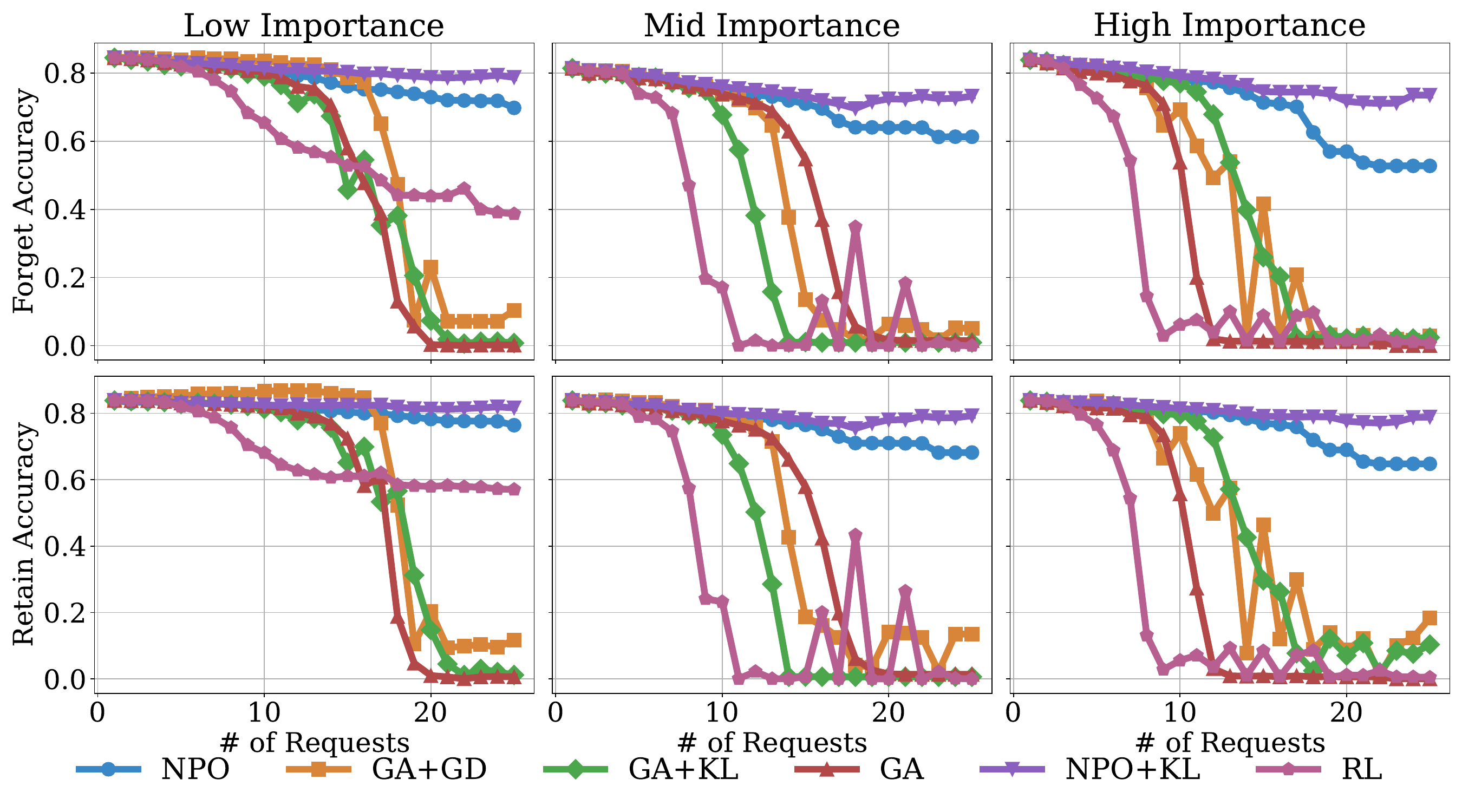}
    \caption{Performance of unlearning methods across importance levels: rows show forget and retain accuracy curves, and columns correspond to low, medium, and high importance. The goal is to select methods that achieve low forget accuracy and high retain accuracy.}
    \label{fig:importance-unlearning}
    \vspace{-10pt}
\end{wrapfigure}
Applying a ``static'' unlearning algorithm across sequential requests induces persistently aligned update directions, compounding gradient instability and parameter drift (Appendix~\ref{impor}).
Dynamically switching algorithms mitigates this by calibrating the update strength and direction to the specific characteristics of each request. 
Since deletion efficacy correlates strongly with sample memorization~\cite{nips/ZhaoKBTT24}, algorithm selection should ideally reflect the memorization level of the incoming data.
However, explicitly computing memorization scores requires prohibitive forward-backward evaluations, necessitating a lightweight proxy.

To enable efficient, request-level adaptation, we introduce \texttt{IMP}, an importance score inspired by gradient-based attribution~\cite{iclr/AnconaCO018}.  This approach enables efficient request-level adaptation without the overhead of memorization-based metrics.
For a filtered request $\mathcal{D}_f^{(t+1)*}$, we compute the $L_2$ norm of the loss gradient with respect to its input embedding:
\begin{equation}
\texttt{IMP}\!\left(L(\mathcal{D}_f^{(t+1)*})\right)=\left\|\nabla_{E(\mathcal{D}_f^{(t+1)*})}L(\mathcal{D}_f^{(t+1)*})\right\|_2,
\end{equation}
where $E(\cdot)$ denotes the embedding function. By quantifying the sensitivity of the loss to the request's embedding, \texttt{IMP} serves as a computationally tractable surrogate for the model's reliance on that data.

Following~\cite{nips/ZhaoKBTT24}, we discretize \texttt{IMP} into three tiers (low, medium, high) to dynamically route the request to one of six standard single-shot unlearning primitives (Section~\ref{related work}). 
Our theoretical analysis (Appendix~\ref{impor}) and empirical findings (Figure~\ref{fig:importance-unlearning}) confirm that calibrating the objective strength according to the \texttt{IMP} score leads to a better forgetting-utility trade-off.

Specifically, low-\texttt{IMP} requests tolerate aggressive algorithms (\eg, RLabel) to maximize forgetting with minimal utility penalty,
medium-\texttt{IMP} requests align with moderated methods (\eg, NPO), and high-\texttt{IMP} requests strictly require conservative approaches (\eg, NPO+KL) to prevent severe utility degradation.
Consequently, dynamically coupling the update magnitude to request influence substantially outperforms fixed-policy unlearning frameworks.

\begin{figure*}[!t]
    \centering
    \includegraphics[width=\textwidth]{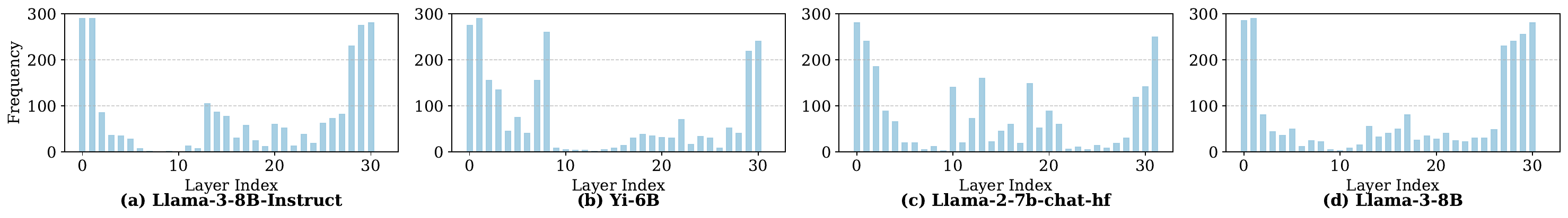}
    \caption{Histogram of layer selection, showing the frequency of each chosen layer}
    \label{fig:layer_histogram}
    \vspace{-5pt}
\end{figure*}


\subsection{Targeted Layer Attribution}
\label{subsec:layer_attribution}
\begin{wrapfigure}{r}{0.36\textwidth}
    \vspace{-12pt}
    \centering
    \includegraphics[width=0.32\textwidth]{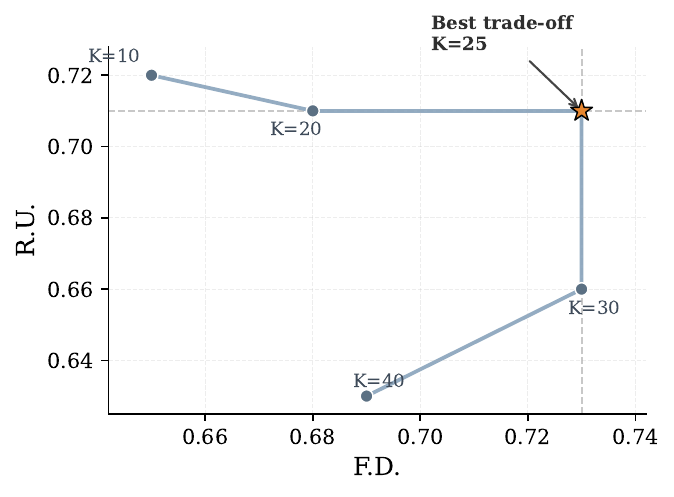}
    \caption{F.D.vs. R.U. under different $K$ values on Llama-3-8B}
    \label{fig:k_tradeoff}
    \vspace{-12pt}
\end{wrapfigure}

Continual unlearning requires a delicate balance: 
updating all layers incurs high computational overhead and degrades model utility, whereas overly sparse updates leave the model vulnerable to post-unlearning knowledge recovery. 
Existing selective strategies fall short in this dynamic setting.
Static interventions, such as freezing bottom layers~\cite{iclr/ZhengCQ025} or exclusively updating last few layers~\cite{corr/goel2022towards}, fail to account for the distributed nature of memorized information. 
Similarly, low-rank-adapter methods (\eg, \(O^3\)~\cite{iclr/gao25}) keep the backbone weights unchanged, allowing forget-set information to remain in the base model and potentially be reactivated~\cite{iclr/0001FWS25}. Further, localized model-editing methods (\eg, AlphaEdit~\cite{iclr/FangJWMSW0C25} and PISCES~\cite{emnlp/gur2025}) restrict updates to a fixed or small parameter subset, implicitly assuming that target knowledge is spatially localized. By limiting updates to rigid subsets, these methods can leave residual knowledge outside the updated subspace, making them also susceptible to knowledge reactivation (Appendix~\ref{layer}).

To ensure robust erasure, unlearning must target a sufficiently broad, request-dependent parameter subset. 
Crucially, enforcing a static layer count fails to scale across diverse architectures: it may be excessive for shallow networks yet entirely insufficient for deeper LLMs, where target knowledge spans a larger topological footprint. 
Therefore, \fit introduces a \textbf{size-adaptive, ratio-based} update rule.
For a model with $L$ layers, we dynamically rank all layers by their request-specific attribution scores and update only the top-$K\%$ (\ie, the \(\lceil K L / 100 \rceil\) highest-scoring layers).
As $K\%$ defines a proportional ratio rather than a fixed ``budget,'' the update capacity naturally adapts to the model's total depth while the specific layers selected remain entirely request-dependent.

\textbf{Layer Selection.}
To identify the target subset $\mathcal{S}_{\text{top-}K\%}$, we estimate each layer's contribution to the unlearning request (Algorithm~\ref{alg:layer-attribution}, Appendix~\ref{layer}). 
Exact Shapley computation over all layer coalitions is exponential in LLMs.
We instead adopt a lightweight leave-one-out approximation~\cite{ijcai/RozemberczkiWBY22}: for each layer $\ell$, we temporarily mask its parameters and measure the resulting loss deviation:

\begin{equation}
    s_\ell = \left| L_{\text{mask}}^{(\ell)} - L_{\text{orig}} \right|,
\end{equation}
where $L_{\text{orig}}$ denotes the standard cross-entropy loss with all activated layers and $L_{\text{mask}}^{(\ell)}$ is the loss evaluated after zeroing out layer $\ell$.
We rank the layers by $s_\ell$ and restrict parameter updates exclusively to the multi-layer perceptron (MLP) and multi-head attention (MHA) modules of the top $K\%$, freezing all other parameters. 
This isolates updates to the components with the highest functional relevance to the forget set, minimizing unnecessary parameter drift (Appendix~\ref{layer}). 


\textbf{Choosing $K\%$.}
Empirical analysis across diverse architectures (Llama-2-7b-chat-hf, Llama-3-8B, Llama-3-8B-Instruct, and Yi-6B) reveals that unlearning attribution consistently concentrates within compact, request-specific modular regions rather than dispersing uniformly across the network (see Figure~\ref{fig:layer_histogram}). 
Through rigorous ablation of update ratios (Figure~\ref{fig:k_tradeoff}), we establish $K = 25\%$ as the optimal threshold for balancing the forgetting-utility trade-off. By establishing a $25\%$ ratio rather than a static layer count, \fit provides a scalable mechanism that automatically adapts to varying model sizes while preserving dynamic, request-dependent layer targeting.



\section{PCH: A Unified Benchmark}{\label{dataset}}




\subsection{The PCH Dataset}
\begin{wraptable}{r}{0.52\textwidth}
    \centering
    \caption{
    Overview of existing benchmarks, where
    ``GPT \& Human'' indicates datasets constructed from GPT-generated candidates that are subsequently verified or refined by human annotators
    }
    \label{tab:dataset-overview}
    \resizebox{0.5\textwidth}{!}{
    \begin{tabular}{llcl}
    \toprule
    \textbf{Dataset} & \textbf{Data Types} & \textbf{Retain Model} & \textbf{Method} \\
    \midrule
    MUSE~\cite{iclr/shi24}             & Copyright            & \checkmark & GPT\&Human \\
    TOFU~\cite{colm/Maini24}           & Personal information & \checkmark & GPT\&Human \\
    WMDP~\cite{icml/LiPGYBGLDGMHLJL24} & Harmful content      & \ding{55}  & Human      \\
    WPU~\cite{emnlp/LiuZJC24}          & Copyright            & \ding{55}  & GPT\&Human \\
    RWKU~\cite{nips/JinCWHYL00024}     & Personal information & \ding{55}  & GPT\&Human \\
    \bottomrule
    \end{tabular}
    }
    \vspace{-10pt}
\end{wraptable}
\emph{Limitations of current datasets.}

Existing unlearning corpora primarily target single-type settings, leaving continual unlearning unexplored~\cite{iclr/YuanPDC0L25}.
TOFU is synthetic (GPT-4 generated with human filtering to enhance diversity and reduce pretraining leakage) and provides a retain model for controlled evaluation~\cite{colm/Maini24}. 
MUSE targets copyright-related deletion with realistic large-scale news and book text, offering structured forget/retain/holdout splits~\cite{iclr/shi24}. 
WMDP focuses on hazardous knowledge removal, with examples manually constructed by experts~\cite{icml/LiPGYBGLDGMHLJL24}. 
As summarized in Table~\ref{tab:dataset-overview}, each dataset centers on a single deletion type: TOFU and RWKU on personal information, MUSE and WPU on copyright violations, and WMDP on harmful content. 
Such type-specific designs may encourage methods to overfit to particular data types, while overlooking data variation within real-world deletion requests. Thus, a unified benchmark covering diverse deletion types is needed to better evaluate continual unlearning in realistic settings.


We introduce \textbf{PCH}, a unified dataset spanning \textbf{P}ersonal information, \textbf{C}opyright, and \textbf{H}armful content, explicitly designed for continual unlearning (see Appendix~\ref{Analysis of PCH} and Figure~\ref{fig:PCH_datatset}). All samples are generated by GPT-4o and then verified by category consistency and basic distributional properties (\eg, text length and token-frequency statistics). Here, ``harmful content'' refers to semantically harmful text (\eg, rumors) rather than genuinely unsafe material~\cite{corr/Hurst2024GPT4oSC}.

Each category contains $200$ samples. 
The entire set of $600$ instances is randomly split into forget and retain subsets. 
Prompts enforce the constraint ``avoid using pre-trained datasets'' to reduce overlap with common pretraining corpora, ensuring retain examples are unseen by the base model. 
As shown in Figure~\ref{fig:retain-forget-llama2} (Appendix~\ref{Analysis of PCH}), a model fine-tuned on the retain set begins with low accuracy but improves steadily, confirming that \textbf{PCH} is out-of-distribution. 
To evaluate forgetting and utility, each instance is converted into a question–answer (QA) pair (Appendix~\ref{sec:qa-pair-construction}, Table~\ref{tab:qa-pairs}).

\subsection{Symmetric Metrics for Forgetting and Utility}
\emph{Limitations of current metrics.}
Only TOFU and MUSE release a retain model, enabling direct comparison between an unlearned model and its original counterpart. 
MUSE reports four distinct metrics: verbatim memorization, knowledge memorization, privacy leakage, and utility preservation, without a single aggregate score. 
This heterogeneity can overweight individual metrics and obscure overall performance, underscoring the need for an integrated measure. 
TOFU combines Forget Quality and Model Utility but i) relies on paraphrase-based answers introducing bias, and ii) applies different evaluation protocols to the forget and retain splits, yielding inconsistent results~\cite{colm/Maini24}.

For remedy, \textbf{PCH} is deliberately constructed to be out-of-distribution to pre-training data, so that a model fine-tuned on $\mathsf{D}_r$ alone serves as the retain model, while a model fine-tuned on $\mathsf{D}_f \cup \mathsf{D}_r$ serves as the fine-tuned model.
We avoid TOFU's costly paraphrasing and adopt three lightweight base metrics: Probability, ROUGE-L, and token-level Accuracy; see Appendix~\ref{app:base-metrics}.
Each metric is applied \emph{identically} to the forget and retain sets, enabling consistent measurement of forgetting and utility.

\textbf{Our aggregated metrics.}
Unlearning evaluation benefits from a single statistic that captures forget-retain trade-offs without being dominated by any one component. 
Following a similar philosophy to TOFU's normalized aggregation~\cite{colm/Maini24}, we derive two symmetric quantities from the three base metrics. 
For the forget and retain sets, we compute the \emph{geometric mean}: 
\begin{align}
F = ( \operatorname{Prob}_{\text{Forget}} \cdot \operatorname{ROUGE}_{\text{Forget}} \cdot \operatorname{Acc}_{\text{Forget}} )^{1/3},
R = ( \operatorname{Prob}_{\text{Retain}} \cdot \operatorname{ROUGE}_{\text{Retain}} \cdot \operatorname{Acc}_{\text{Retain}} )^{1/3},
\end{align}
and similarly obtain $FQ$ and $RQ$ for retain model.
The geometric mean balances improvements and prevents any metric from dominating the result. Based on them, we define two regularized measures:
\begin{align}
\operatorname{F.D.} = \max\left( 0,\, 1 - \left| F / FQ - 1 \right| \right),
\operatorname{R.U.} = \max\left( 0,\, 1 - \left| R / RQ - 1 \right| \right).
\end{align}
Forget Degree (F.D.) and Retain Utility (R.U.) measure alignment to the retain model on the forget and retain sets, respectively. Since the retain model approximates retraining, the goal is closer retain-model alignment, rather than assuming ``larger $R$ / smaller $F$ is always better.''
Figure~\ref{fig:fd-geo-surface} (Appendix~\ref{app:base-metrics}) illustrates these properties.
Panels (a--c) plot the geometric mean of two metrics with the third fixed at $0.7$; the sharp decline from one weak component shows that it penalizes imbalance, making it a balanced aggregator.
Panel (d) shows F.D.\ as a function of $F/FQ$, with a symmetric, approximately linear drop from the optimum, making F.D.\ scale-invariant and interpretable.

\begin{table}[!t]
\centering
\caption{Continual unlearning performance on Llama-3-8B and Qwen3-14B across request stages}
\label{tab:main_results_part1}
\footnotesize
\setlength{\tabcolsep}{2.8pt}
\renewcommand{\arraystretch}{1.08}
\resizebox{\textwidth}{!}{
\begin{tabular}{lccccccccccccc}
\toprule
\rowcolor{headergray}
\textbf{Method}
& \multicolumn{2}{c}{\textbf{60 req}} 
& \multicolumn{2}{c}{\textbf{120 req}} 
& \multicolumn{2}{c}{\textbf{180 req}} 
& \multicolumn{2}{c}{\textbf{240 req}} 
& \multicolumn{5}{c}{\textbf{300 req}} \\
\cmidrule(lr){2-3}
\cmidrule(lr){4-5}
\cmidrule(lr){6-7}
\cmidrule(lr){8-9}
\cmidrule(lr){10-14}
\rowcolor{subheadergray}
& \textbf{F.D.$\uparrow$} & \textbf{R.U.$\uparrow$} 
& \textbf{F.D.$\uparrow$} & \textbf{R.U.$\uparrow$} 
& \textbf{F.D.$\uparrow$} & \textbf{R.U.$\uparrow$} 
& \textbf{F.D.$\uparrow$} & \textbf{R.U.$\uparrow$} 
& \textbf{F.D.$\uparrow$} & \textbf{R.U.$\uparrow$} 
& \textbf{MMLU$\uparrow$} & \textbf{CSQA$\uparrow$} & \textbf{GSM8K$\uparrow$} \\
\midrule

\rowcolor{modelgray}
\multicolumn{14}{c}{\textbf{Llama-3-8B}} \\

\rowcolor{retainlight}
\textbf{Retain Model}
& \scorestd{1.00}{0.00} & \scorestd{1.00}{0.00}
& \scorestd{1.00}{0.00} & \scorestd{1.00}{0.00}
& \scorestd{1.00}{0.00} & \scorestd{1.00}{0.00}
& \scorestd{1.00}{0.00} & \scorestd{1.00}{0.00}
& \scorestd{1.00}{0.00} & \scorestd{1.00}{0.00}
& \scorestd{65.97}{0.00} & \scorestd{70.02}{0.00} & \scorestd{58.83}{0.00} \\

GA      
& \scorestd{0.00}{0.00} & \scorestd{0.00}{0.00}
& \scorestd{0.00}{0.00} & \scorestd{0.00}{0.00}
& \scorestd{0.00}{0.00} & \scorestd{0.00}{0.00}
& \scorestd{0.00}{0.00} & \scorestd{0.00}{0.00}
& \scorestd{0.00}{0.00} & \scorestd{0.00}{0.00}
& \scorestd{24.42}{1.56} & \scorestd{19.57}{1.58} & \scorestd{0.00}{0.00} \\

GA+GD   
& \scorestd{0.01}{0.04} & \scorestd{0.01}{0.03}
& \scorestd{0.06}{0.05} & \scorestd{0.08}{0.05}
& \scorestd{0.17}{0.06} & \scorestd{0.14}{0.05}
& \scorestd{0.46}{0.07} & \scorestd{0.44}{0.07}
& \scorestd{0.55}{0.08} & \scorestd{0.48}{0.07}
& \scorestd{62.17}{0.98} & \scorestd{68.14}{0.86} & \scorestd{47.08}{1.14} \\

GA+KL   
& \scorestd{0.00}{0.00} & \scorestd{0.00}{0.00}
& \scorestd{0.00}{0.00} & \scorestd{0.00}{0.00}
& \scorestd{0.00}{0.00} & \scorestd{0.00}{0.00}
& \scorestd{0.00}{0.00} & \scorestd{0.00}{0.00}
& \scorestd{0.00}{0.00} & \scorestd{0.00}{0.00}
& \scorestd{24.46}{1.54} & \scorestd{20.72}{1.50} & \scorestd{0.00}{0.00} \\

NPO     
& \scorestd{0.25}{0.06} & \scorestd{0.28}{0.06}
& \scorestd{0.22}{0.06} & \scorestd{0.20}{0.05}
& \scorestd{0.15}{0.05} & \scorestd{0.13}{0.05}
& \scorestd{0.12}{0.05} & \scorestd{0.10}{0.05}
& \scorestd{0.12}{0.06} & \scorestd{0.09}{0.05}
& \scorestd{58.35}{1.08} & \scorestd{58.49}{1.16} & \scorestd{50.81}{1.02} \\

NPO+KL  
& \scorestd{0.66}{0.07} & \scorestd{0.68}{0.06}
& \scorestd{0.57}{0.07} & \scorestd{0.58}{0.06}
& \scorestd{0.64}{0.06} & \scorestd{0.65}{0.06}
& \scorestd{0.70}{0.07} & \scorestd{0.66}{0.06}
& \scorestd{0.59}{0.07} & \scorestd{0.60}{0.06}
& \scorestd{61.18}{0.96} & \scorestd{61.03}{1.05} & \scorestd{52.99}{0.94} \\

RLabel  
& \scorestd{0.01}{0.04} & \scorestd{0.01}{0.03}
& \scorestd{0.01}{0.04} & \scorestd{0.00}{0.00}
& \scorestd{0.01}{0.04} & \scorestd{0.01}{0.04}
& \scorestd{0.00}{0.00} & \scorestd{0.00}{0.00}
& \scorestd{0.00}{0.00} & \scorestd{0.00}{0.00}
& \scorestd{24.76}{1.52} & \scorestd{19.57}{1.56} & \scorestd{0.00}{0.00} \\

PISCES  
& \scorestd{0.00}{0.00} & \scorestd{0.00}{0.00}
& \scorestd{0.00}{0.00} & \scorestd{0.00}{0.00}
& \scorestd{0.00}{0.00} & \scorestd{0.00}{0.00}
& \scorestd{0.00}{0.00} & \scorestd{0.00}{0.00}
& \scorestd{0.00}{0.00} & \scorestd{0.00}{0.00}
& \scorestd{25.58}{1.48} & \scorestd{21.58}{1.46} & \scorestd{0.00}{0.00} \\

ALKN    
& \scorestd{0.84}{0.05} & \scorestd{0.80}{0.05}
& \scorestd{0.78}{0.06} & \scorestd{\textbf{0.76}}{0.05}
& \scorestd{0.74}{0.06} & \scorestd{0.69}{0.06}
& \scorestd{0.71}{0.07} & \scorestd{0.65}{0.06}
& \scorestd{0.60}{0.07} & \scorestd{0.62}{0.06}
& \scorestd{62.17}{0.88} & \scorestd{69.54}{0.42} & \bestscorestd{57.09}{0.92} \\

$O^3$      
& \scorestd{\textbf{0.93}}{0.04} & \scorestd{\textbf{0.90}}{0.04}
& \scorestd{\textbf{0.82}}{0.05} & \scorestd{0.75}{0.05}
& \scorestd{0.73}{0.06} & \scorestd{0.68}{0.06}
& \scorestd{0.69}{0.07} & \scorestd{0.62}{0.06}
& \scorestd{0.64}{0.07} & \scorestd{0.63}{0.06}
& -- & -- & -- \\

\rowcolor{ourslight}
\textbf{Ours}    
& \scorestd{0.90}{0.04} & \scorestd{0.88}{0.04}
& \scorestd{0.75}{0.05} & \scorestd{0.71}{0.05}
& \bestscorestd{0.75}{0.05} & \bestscorestd{0.72}{0.05}
& \bestscorestd{0.76}{0.05} & \bestscorestd{0.70}{0.05}
& \bestscorestd{0.73}{0.05} & \bestscorestd{0.71}{0.05}
& \bestscorestd{65.56}{0.34} & \bestscorestd{70.01}{0.01} & \scorestd{57.02}{0.86} \\

\midrule

\rowcolor{modelgray}
\multicolumn{14}{c}{\textbf{Qwen3-14B}} \\

\rowcolor{retainlight}
\textbf{Retain Model}
& \scorestd{1.00}{0.00} & \scorestd{1.00}{0.00}
& \scorestd{1.00}{0.00} & \scorestd{1.00}{0.00}
& \scorestd{1.00}{0.00} & \scorestd{1.00}{0.00}
& \scorestd{1.00}{0.00} & \scorestd{1.00}{0.00}
& \scorestd{1.00}{0.00} & \scorestd{1.00}{0.00}
& \scorestd{79.99}{0.00} & \scorestd{81.33}{0.00} & \scorestd{84.84}{0.00} \\

GA      
& \scorestd{0.62}{0.08} & \scorestd{0.68}{0.07}
& \scorestd{0.27}{0.07} & \scorestd{0.27}{0.06}
& \scorestd{0.00}{0.00} & \scorestd{0.00}{0.00}
& \scorestd{0.00}{0.00} & \scorestd{0.00}{0.00}
& \scorestd{0.00}{0.00} & \scorestd{0.00}{0.00}
& \scorestd{45.29}{1.56} & \scorestd{25.14}{1.62} & \scorestd{0.00}{0.00} \\

GA+GD   
& \scorestd{0.87}{0.06} & \scorestd{0.85}{0.06}
& \scorestd{0.75}{0.07} & \scorestd{0.61}{0.07}
& \scorestd{0.64}{0.08} & \scorestd{0.62}{0.07}
& \scorestd{0.69}{0.08} & \scorestd{0.65}{0.07}
& \scorestd{0.52}{0.08} & \scorestd{0.44}{0.07}
& \scorestd{72.35}{1.04} & \scorestd{77.64}{0.88} & \scorestd{69.30}{1.16} \\

GA+KL   
& \scorestd{0.51}{0.07} & \scorestd{0.55}{0.07}
& \scorestd{0.28}{0.07} & \scorestd{0.29}{0.06}
& \scorestd{0.06}{0.05} & \scorestd{0.06}{0.05}
& \scorestd{0.00}{0.00} & \scorestd{0.00}{0.00}
& \scorestd{0.00}{0.00} & \scorestd{0.00}{0.00}
& \scorestd{68.08}{1.22} & \scorestd{43.49}{1.46} & \scorestd{0.00}{0.00} \\

NPO     
& \scorestd{0.69}{0.08} & \scorestd{0.75}{0.07}
& \scorestd{0.55}{0.07} & \scorestd{0.58}{0.07}
& \scorestd{0.49}{0.08} & \scorestd{0.48}{0.07}
& \scorestd{0.50}{0.08} & \scorestd{0.43}{0.07}
& \scorestd{0.48}{0.08} & \scorestd{0.40}{0.07}
& \scorestd{72.14}{1.08} & \scorestd{72.48}{1.12} & \scorestd{57.85}{1.36} \\

NPO+KL  
& \scorestd{0.69}{0.07} & \scorestd{0.74}{0.07}
& \scorestd{0.63}{0.07} & \scorestd{0.68}{0.06}
& \scorestd{0.64}{0.07} & \scorestd{0.65}{0.06}
& \scorestd{0.70}{0.07} & \scorestd{0.63}{0.06}
& \scorestd{0.69}{0.07} & \scorestd{0.60}{0.06}
& \scorestd{76.52}{0.82} & \scorestd{75.36}{1.02} & \scorestd{73.14}{0.94} \\

RLabel  
& \scorestd{0.25}{0.06} & \scorestd{0.29}{0.06}
& \scorestd{0.09}{0.05} & \scorestd{0.09}{0.05}
& \scorestd{0.05}{0.05} & \scorestd{0.05}{0.05}
& \scorestd{0.05}{0.05} & \scorestd{0.04}{0.04}
& \scorestd{0.04}{0.05} & \scorestd{0.03}{0.04}
& \scorestd{47.49}{1.52} & \scorestd{28.55}{1.58} & \scorestd{0.00}{0.00} \\

PISCES  
& \scorestd{0.00}{0.00} & \scorestd{0.00}{0.00}
& \scorestd{0.00}{0.00} & \scorestd{0.00}{0.00}
& \scorestd{0.00}{0.00} & \scorestd{0.00}{0.00}
& \scorestd{0.00}{0.00} & \scorestd{0.00}{0.00}
& \scorestd{0.00}{0.00} & \scorestd{0.00}{0.00}
& \scorestd{54.68}{1.42} & \scorestd{35.49}{1.54} & \scorestd{34.59}{1.50} \\

ALKN    
& \scorestd{0.84}{0.06} & \scorestd{0.80}{0.06}
& \scorestd{\textbf{0.78}}{0.06} & \scorestd{0.79}{0.06}
& \scorestd{\textbf{0.77}}{0.06} & \scorestd{\textbf{0.75}}{0.05}
& \scorestd{0.75}{0.06} & \scorestd{\textbf{0.77}}{0.05}
& \scorestd{0.74}{0.06} & \scorestd{0.71}{0.05}
& \scorestd{77.40}{0.64} & \bestscorestd{78.26}{0.72} & \bestscorestd{74.68}{0.78} \\

$O^3$      
& \scorestd{0.87}{0.05} & \scorestd{\textbf{0.91}}{0.04}
& \scorestd{0.74}{0.06} & \scorestd{0.73}{0.05}
& \scorestd{0.70}{0.06} & \scorestd{0.71}{0.05}
& \scorestd{0.75}{0.06} & \scorestd{0.73}{0.05}
& \scorestd{0.86}{0.06} & \scorestd{0.75}{0.05}
& -- & -- & -- \\

\rowcolor{ourslight}
\textbf{Ours}    
& \bestscorestd{0.89}{0.04} & \scorestd{0.90}{0.04}
& \scorestd{0.75}{0.05} & \scorestd{0.72}{0.05}
& \scorestd{0.76}{0.05} & \scorestd{0.72}{0.05}
& \bestscorestd{0.83}{0.05} & \scorestd{0.73}{0.05}
& \bestscorestd{0.89}{0.05} & \bestscorestd{0.78}{0.04}
& \bestscorestd{79.15}{0.54} & \scorestd{77.31}{0.86} & \scorestd{74.28}{0.82} \\

\bottomrule
\end{tabular}
}
\vspace{-2pt}
\end{table}

\begin{table}[!t]
\centering
\caption{Ablation results on Llama-3-8B across request stages}
\label{tab:ablation_llama3_8b}
\footnotesize
\setlength{\tabcolsep}{3.2pt}
\renewcommand{\arraystretch}{1.08}
\resizebox{\textwidth}{!}{
\begin{tabular}{lccccccccccccc}
\toprule
\rowcolor{headergray}
\textbf{Method}
& \multicolumn{2}{c}{\textbf{60 req}} 
& \multicolumn{2}{c}{\textbf{120 req}} 
& \multicolumn{2}{c}{\textbf{180 req}} 
& \multicolumn{2}{c}{\textbf{240 req}} 
& \multicolumn{5}{c}{\textbf{300 req}} \\
\cmidrule(lr){2-3}
\cmidrule(lr){4-5}
\cmidrule(lr){6-7}
\cmidrule(lr){8-9}
\cmidrule(lr){10-14}
\rowcolor{subheadergray}
& \textbf{F.D.$\uparrow$} & \textbf{R.U.$\uparrow$} 
& \textbf{F.D.$\uparrow$} & \textbf{R.U.$\uparrow$} 
& \textbf{F.D.$\uparrow$} & \textbf{R.U.$\uparrow$} 
& \textbf{F.D.$\uparrow$} & \textbf{R.U.$\uparrow$} 
& \textbf{F.D.$\uparrow$} & \textbf{R.U.$\uparrow$} 
& \textbf{MMLU$\uparrow$} & \textbf{CSQA$\uparrow$} & \textbf{GSM8K$\uparrow$} \\
\midrule

\rowcolor{modelgray}
\multicolumn{14}{c}{\textbf{Llama-3-8B}} \\

\rowcolor{retainlight}
\textbf{Retain Model}
& \scorestd{1.00}{0.00} & \scorestd{1.00}{0.00}
& \scorestd{1.00}{0.00} & \scorestd{1.00}{0.00}
& \scorestd{1.00}{0.00} & \scorestd{1.00}{0.00}
& \scorestd{1.00}{0.00} & \scorestd{1.00}{0.00}
& \scorestd{1.00}{0.00} & \scorestd{1.00}{0.00}
& \scorestd{65.97}{0.00} & \scorestd{70.02}{0.00} & \scorestd{58.83}{0.00} \\

w/o adaptive algorithm
& \scorestd{0.79}{0.05} & \scorestd{0.82}{0.05}
& \scorestd{0.50}{0.06} & \scorestd{0.49}{0.05}
& \scorestd{0.16}{0.06} & \scorestd{0.10}{0.05}
& \scorestd{0.20}{0.06} & \scorestd{0.14}{0.05}
& \scorestd{0.14}{0.07} & \scorestd{0.07}{0.06}
& \scorestd{62.40}{0.96} & \scorestd{63.80}{1.08} & \scorestd{14.12}{1.42} \\

w/o filtering
& \scorestd{0.65}{0.07} & \scorestd{0.76}{0.06}
& \scorestd{0.59}{0.07} & \scorestd{0.63}{0.06}
& \scorestd{0.61}{0.06} & \scorestd{0.61}{0.06}
& \scorestd{0.64}{0.07} & \scorestd{0.59}{0.06}
& \scorestd{0.66}{0.07} & \scorestd{0.63}{0.06}
& \scorestd{64.17}{0.72} & \scorestd{68.21}{0.84} & \scorestd{55.72}{1.02} \\

w/o chosen layers
& \scorestd{0.53}{0.06} & \scorestd{0.59}{0.05}
& \scorestd{0.43}{0.07} & \scorestd{0.51}{0.06}
& \scorestd{0.43}{0.06} & \scorestd{0.51}{0.05}
& \scorestd{0.51}{0.05} & \scorestd{0.53}{0.06}
& \scorestd{0.40}{0.07} & \scorestd{0.48}{0.06}
& \scorestd{64.02}{0.78} & \scorestd{67.12}{0.92} & \scorestd{54.98}{1.08} \\

\rowcolor{ourslight}
\textbf{Ours}
& \bestscorestd{0.90}{0.04} & \bestscorestd{0.88}{0.04}
& \bestscorestd{0.75}{0.05} & \bestscorestd{0.71}{0.05}
& \bestscorestd{0.75}{0.05} & \bestscorestd{0.72}{0.05}
& \bestscorestd{0.76}{0.05} & \bestscorestd{0.70}{0.05}
& \bestscorestd{0.73}{0.05} & \bestscorestd{0.71}{0.05}
& \bestscorestd{65.56}{0.34} & \bestscorestd{70.01}{0.01} & \bestscorestd{57.02}{0.86} \\

\bottomrule
\end{tabular}
}
\vspace{-2pt}
\end{table}    

\section{Experiment}{\label{experi}}

\subsection{Experimental Setup}

\textbf{Dataset and Models.} All experiments are run on \textbf{PCH} benchmark;
We evaluate \fit on five widely used open-source LLMs
: 
Yi-6B~\cite{corr/Young24},
Llama-2-7b-chat-hf~\cite{corr/Touvro23}, 
Llama-3-8B~\cite{corr/Dubey24}, 
Llama-3-8B-Instruct~\cite{corr/Dubey24}, 
and Qwen3-14B~\cite{corr/Yang25}.
Since the pre-training corpus $\mathsf{D}$ is unavailable, retraining on $\mathsf{D}_r$ is infeasible.
As outlined in Section~\ref{problem} and Appendix~\ref{Fine-tuned and Retain models}, we instead construct synthesis ``$\mathsf{D}_f$'' and ``$\mathsf{D}_r$,'' from which we derive a \emph{fine-tuned model} and a \emph{retain model} as proxies.

\textbf{Baselines and Evaluation Metrics.}
We evaluate our framework against several unlearning algorithms:
GA, GA+GD, GA+KL, NPO, NPO+KL, RLabel, 
PISCES~\cite{emnlp/gur2025}, $O^3$~\cite{iclr/gao25}, and ALKN~\cite{icml/wuerkaixi2025adaptive}; We use the two metrics, F.D. and R.U., to quantify the forgetting effectiveness and utility preservation.
Final model downstream performance is reported on MMLU~\cite{iclr/HendrycksBBZMSS21}, CommonsenseQA (CSQA)~\cite{naacl/TalmorHLB19}, and GSM8K~\cite{corr/cobbe21}.
Post-unlearning recovery is evaluated under two settings: i) \emph{Relearning via fine-tuning}~\cite{icml/xu2025}, where the unlearned model is fine-tuned on mixed retain/unrelated data, retain-only data, or unrelated-only data. We do not assume access to the forget set, as this is unrealistic and undermines the privacy goal. ii) \emph{quantization attacks}~\cite{iclr/ZhangWLWTL00W25}, where model weights are compressed to \texttt{int4} to test whether forgotten knowledge can re-emerge.
Implementation details are provided in Appendix~\ref{Experimental configuration details}.




\subsection{Results}
\paragraph{Forgetting-Utility Trade-off.}
Table~\ref{tab:main_results_part1} reports F.D., R.U., and downstream performance. Overall, \fit achieves a strong, stable trade-off on Llama-3-8B and scales to Qwen3-14B, with small deviations. Methods like NPO+KL, \(O^3\), and ALKN sometimes yield higher R.U. but lower F.D., indicating incomplete forgetting; aggressive baselines like GA and RLabel degrade quickly. After 300 requests, \fit preserves strong downstream performance, achieving best MMLU and CSQA on Llama-3-8B and best MMLU on Qwen3-14B, while competitive on GSM8K. We exclude \(O^3\) from downstream evaluation because its dynamic OOD detection is incompatible with OpenCompass's fixed protocol~\cite{2023opencompass}. Results on remaining models and efficiency (GPU memory) appear in Appendix~\ref{Additional Result and Efficiency Analysi}.

\paragraph{Post-unlearning Recovery.} \label{sec:robustness}

\begin{wrapfigure}{r}{0.48\textwidth}
    \centering
    \vspace{-12pt}
    \includegraphics[width=0.48\textwidth]{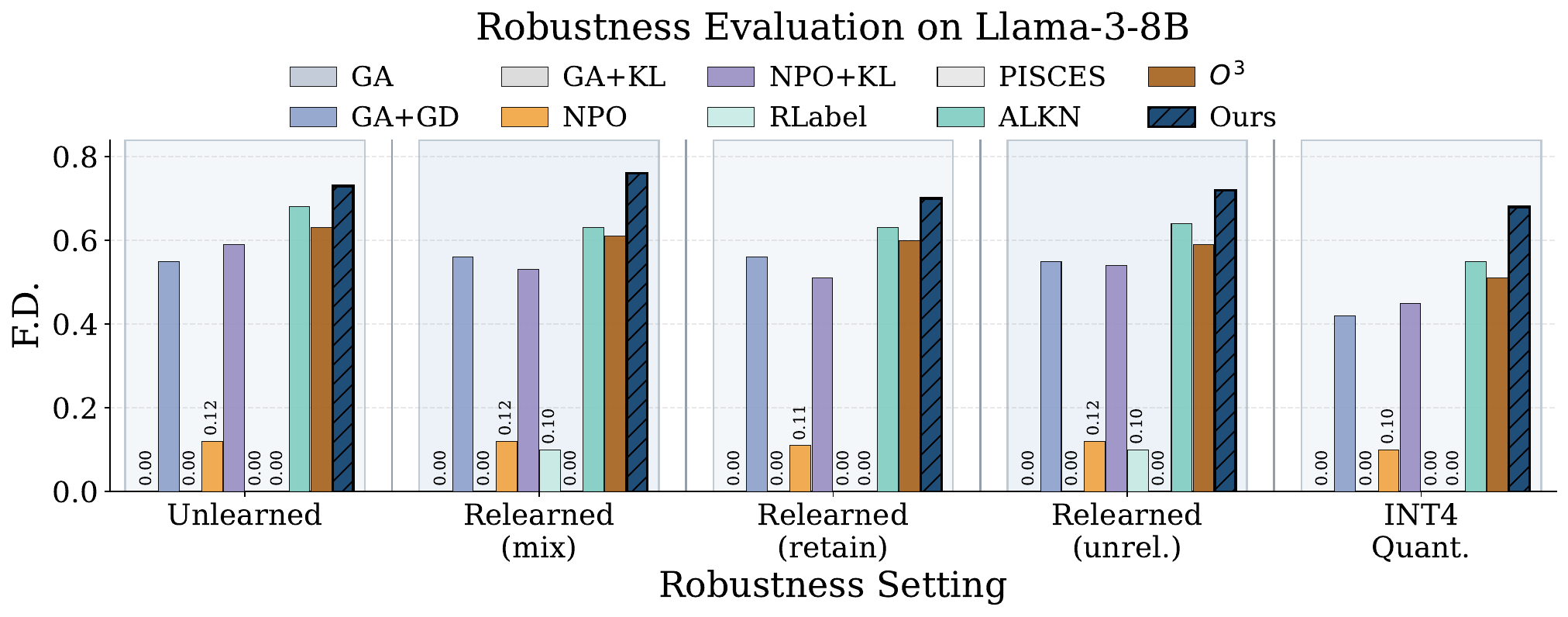}
    \caption{Robustness evaluation under relearning and quantization attacks on Llama-3-8B}
    \label{fig:robustness_llama3_8b}
    \vspace{-13pt}
\end{wrapfigure}
Figure~\ref{fig:robustness_llama3_8b} evaluates post-unlearning recovery on Llama-3-8B under three relearning settings and \texttt{int4} quantization. Across all settings, aggressive baselines such as GA and RLabel remain near zero, while conservative methods such as NPO+KL, ALKN, and \(O^3\) retain moderate F.D. \fit shows the strongest overall robustness, achieving the highest F.D. after mixed-data relearning and quantization (\texttt{int4}) while remaining competitive under retain-only and unrelated-data relearning. This suggests that \fit better resists post-unlearning recovery attempts.



\paragraph{Ablation.} Table~\ref{tab:ablation_llama3_8b} ablates three core components of \fit: adaptive algorithm selection, embedding-based filtering, and targeted layer updates. Removing adaptive algorithm selection causes the largest drop, with F.D. and R.U. decreasing sharply and GSM8K falling to \(14.12\) after 300 requests, highlighting its role against forgetting. Removing filtering yields a milder but consistent decline, suggesting redundancy control stabilizes updates. Excluding targeted layer updates weakens performance, showing request-specific layer selection reduces parameter drift. Overall, the full framework achieves the best F.D.--R.U. trade-off and downstream performance, confirming component complementarity.

\begin{wrapfigure}{r}{0.36\textwidth}
    \vspace{-14pt}
    \centering
    \includegraphics[width=0.35\textwidth]{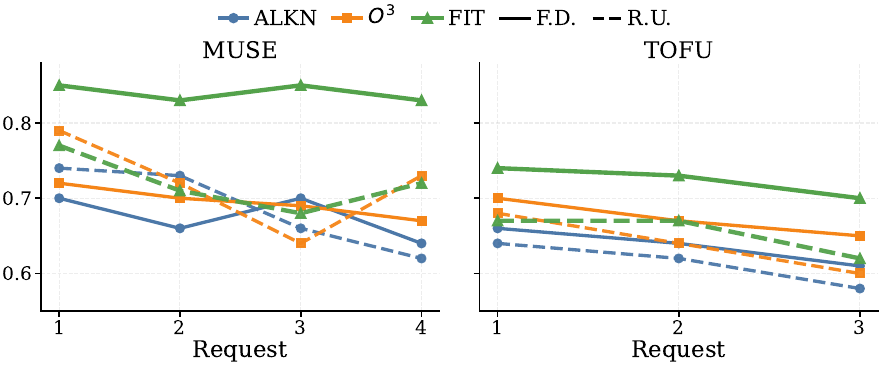}
    \caption{Cross-benchmark evaluation on MUSE and TOFU}
    \label{fig:muse_tofu_llama3_8b}
    \vspace{-20pt}
\end{wrapfigure}

\subsection{Generalization to TOFU and MUSE.}
We further evaluate FIT on MUSE~\cite{iclr/shi24} and TOFU~\cite{colm/Maini24} to assess its generalization beyond \textbf{PCH}, comparing with the state-of-the-art baselines ALKN and $O^3$.
MUSE is a real-world unlearning benchmark from news and book data, while TOFU is a controlled synthetic benchmark for profile-based forget requests.
As shown in Figure~\ref{fig:muse_tofu_llama3_8b}, FIT consistently achieves a stronger F.D./R.U. trade-off than ALKN and $O^3$ across request settings.
These results suggest that FIT is not merely calibrated to \textbf{PCH}, but generalizes across both real-world and other synthetic unlearning benchmarks.

\section{Conclusion}
\label{conclu}

We introduced \fit, a robust and practical framework for continual LLM unlearning under sequential requests. \fit integrates redundancy \underline{F}iltering, \underline{I}mportance-aware algorithm selection, and \underline{T}argeted layer attribution to stabilize sequential updates, reduce redundant gradient accumulation, and limit unnecessary parameter drift. We also proposed \textbf{PCH}, a unified benchmark covering \textbf{P}ersonal information, \textbf{C}opyright, and \textbf{H}armful content, together with F.D. and R.U. to measure the forgetting-utility trade-off. Experiments show that \fit achieves stronger forgetting and higher utility than priors, while preserving downstream performance and resisting relearning and quantization attacks.



\bibliographystyle{plain}
\bibliography{Reference}

@misc{corr/Touvro23,
  author       = {Hugo Touvron and
                  Louis Martin and
                  Kevin Stone and
                  Peter Albert and
                  Amjad Almahairi and
                  Yasmine Babaei and
                  Nikolay Bashlykov and
                  Soumya Batra and
                  Prajjwal Bhargava and
                  Shruti Bhosale and
                  Dan Bikel and
                  Lukas Blecher and
                  Cristian Canton{-}Ferrer and
                  Moya Chen and
                  Guillem Cucurull and
                  David Esiobu and
                  Jude Fernandes and
                  Jeremy Fu and
                  Wenyin Fu and
                  Brian Fuller and
                  Cynthia Gao and
                  Vedanuj Goswami and
                  Naman Goyal and
                  Anthony Hartshorn and
                  Saghar Hosseini and
                  Rui Hou and
                  Hakan Inan and
                  Marcin Kardas and
                  Viktor Kerkez and
                  Madian Khabsa and
                  Isabel Kloumann and
                  Artem Korenev and
                  Punit Singh Koura and
                  Marie{-}Anne Lachaux and
                  Thibaut Lavril and
                  Jenya Lee and
                  Diana Liskovich and
                  Yinghai Lu and
                  Yuning Mao and
                  Xavier Martinet and
                  Todor Mihaylov and
                  Pushkar Mishra and
                  Igor Molybog and
                  Yixin Nie and
                  Andrew Poulton and
                  Jeremy Reizenstein and
                  Rashi Rungta and
                  Kalyan Saladi and
                  Alan Schelten and
                  Ruan Silva and
                  Eric Michael Smith and
                  Ranjan Subramanian and
                  Xiaoqing Ellen Tan and
                  Binh Tang and
                  Ross Taylor and
                  Adina Williams and
                  Jian Xiang Kuan and
                  Puxin Xu and
                  Zheng Yan and
                  Iliyan Zarov and
                  Yuchen Zhang and
                  Angela Fan and
                  Melanie Kambadur and
                  Sharan Narang and
                  Aur{\'{e}}lien Rodriguez and
                  Robert Stojnic and
                  Sergey Edunov and
                  Thomas Scialom},
  title        = {Llama 2: Open Foundation and Fine-Tuned Chat Models},
  howpublished = {arXiv:2307.09288},
  year         = {2023}
}

@inproceedings{nips/MengBAB22,
  author       = {Kevin Meng and
                  David Bau and
                  Alex Andonian and
                  Yonatan Belinkov},
  title        = {Locating and Editing Factual Associations in {GPT}},
  booktitle    = {{NeurIPS}},
  year         = {2022}
}

@inproceedings{emnlp/GevaBFG23,
  author       = {Mor Geva and
                  Jasmijn Bastings and
                  Katja Filippova and
                  Amir Globerson},
  title        = {Dissecting Recall of Factual Associations in Auto-Regressive Language
                  Models},
  booktitle    = {{EMNLP}},
  pages        = {12216--12235},
  year         = {2023}
}

@inproceedings{iclr/HuSWALWWC22,
  author       = {Edward J. Hu and
                  Yelong Shen and
                  Phillip Wallis and
                  Zeyuan Allen{-}Zhu and
                  Yuanzhi Li and
                  Shean Wang and
                  Lu Wang and
                  Weizhu Chen},
  title        = {LoRA: Low-Rank Adaptation of Large Language Models},
  booktitle    = {{ICLR}},
  year         = {2022}
}

@inproceedings{acl/YaoCDNWCY24,
  author       = {Jin Yao and
                  Eli Chien and
                  Minxin Du and
                  Xinyao Niu and
                  Tianhao Wang and
                  Zezhou Cheng and
                  Xiang Yue},
  title        = {Machine Unlearning of Pre-trained Large Language Models},
  booktitle    = {{ACL}},
  pages        = {8403--8419},
  year         = {2024}
}

@inproceedings{acl/JangYYCLLS23,
  author       = {Joel Jang and
                  Dongkeun Yoon and
                  Sohee Yang and
                  Sungmin Cha and
                  Moontae Lee and
                  Lajanugen Logeswaran and
                  Minjoon Seo},
  title        = {Knowledge Unlearning for Mitigating Privacy Risks in Language Models},
  booktitle    = {{ACL}},
  pages        = {14389--14408},
  year         = {2023}
}

@inproceedings{icml/PawelczykNL24,
  author       = {Martin Pawelczyk and
                  Seth Neel and
                  Himabindu Lakkaraju},
  title        = {In-Context Unlearning: Language Models as Few-Shot Unlearners},
  booktitle    = {{ICML}},
  year         = {2024}
}

@inproceedings{icml/LiPGYBGLDGMHLJL24,
  author       = {Nathaniel Li and
                  Alexander Pan and
                  Anjali Gopal and
                  Summer Yue and
                  Daniel Berrios and
                  Alice Gatti and
                  Justin D. Li and
                  Ann{-}Kathrin Dombrowski and
                  Shashwat Goel and
                  Gabriel Mukobi and
                  Nathan Helm{-}Burger and
                  Rassin Lababidi and
                  Lennart Justen and
                  Andrew B. Liu and
                  Michael Chen and
                  Isabelle Barrass and
                  Oliver Zhang and
                  Xiaoyuan Zhu and
                  Rishub Tamirisa and
                  Bhrugu Bharathi and
                  Ariel Herbert{-}Voss and
                  Cort B. Breuer and
                  Andy Zou and
                  Mantas Mazeika and
                  Zifan Wang and
                  Palash Oswal and
                  Weiran Lin and
                  Adam A. Hunt and
                  Justin Tienken{-}Harder and
                  Kevin Y. Shih and
                  Kemper Talley and
                  John Guan and
                  Ian Steneker and
                  David Campbell and
                  Brad Jokubaitis and
                  Steven Basart and
                  Stephen Fitz and
                  Ponnurangam Kumaraguru and
                  Kallol Krishna Karmakar and
                  Uday Kiran Tupakula and
                  Vijay Varadharajan and
                  Yan Shoshitaishvili and
                  Jimmy Ba and
                  Kevin M. Esvelt and
                  Alexandr Wang and
                  Dan Hendrycks},
  title        = {The {WMDP} Benchmark: Measuring and Reducing Malicious Use with Unlearning},
  booktitle    = {{ICML}},
  year         = {2024}
}

@inproceedings{nips/Liu24,
  author       = {Chris Yuhao Liu and
                  Yaxuan Wang and
                  Jeffrey Flanigan and
                  Yang Liu},
  title        = {Large Language Model Unlearning via Embedding-Corrupted Prompts},
  booktitle    = {{NeurIPS}},
  year         = {2024}
}

@inproceedings{sp/BourtouleCCJTZL21,
  author       = {Lucas Bourtoule and
                  Varun Chandrasekaran and
                  Christopher A. Choquette{-}Choo and
                  Hengrui Jia and
                  Adelin Travers and
                  Baiwu Zhang and
                  David Lie and
                  Nicolas Papernot},
  title        = {Machine Unlearning},
  booktitle    = {{S\&P}},
  pages        = {141--159},
  year         = {2021}
}

@inproceedings{iclr/IlharcoRWSHF23,
  author       = {Gabriel Ilharco and
                  Marco T{\'{u}}lio Ribeiro and
                  Mitchell Wortsman and
                  Ludwig Schmidt and
                  Hannaneh Hajishirzi and
                  Ali Farhadi},
  title        = {Editing models with task arithmetic},
  booktitle    = {{ICLR}},
  year         = {2023}
}

@inproceedings{colm/Maini24,
  author       = {Pratyush Maini and
                  Zhili Feng and
                  Avi Schwarzschild and
                  Zachary C. Lipton and
                  J. Zico Kolter},
  title        = {{TOFU:} {A} Task of Fictitious Unlearning for LLMs},
  booktitle    = {{COLM}},
  year         = {2024}
}

@inproceedings{uss/CarliniTWJHLRBS21,
  author       = {Nicholas Carlini and
                  Florian Tram{\`{e}}r and
                  Eric Wallace and
                  Matthew Jagielski and
                  Ariel Herbert{-}Voss and
                  Katherine Lee and
                  Adam Roberts and
                  Tom B. Brown and
                  Dawn Song and
                  {\'{U}}lfar Erlingsson and
                  Alina Oprea and
                  Colin Raffel},
  title        = {Extracting Training Data from Large Language Models},
  booktitle    = {{USENIX} Security},
  pages        = {2633--2650},
  year         = {2021}
}

@misc{corr/Dubey24,
  author       = {Abhimanyu Dubey and
                  Abhinav Jauhri and
                  Abhinav Pandey and
                  Abhishek Kadian and
                  Ahmad Al{-}Dahle and
                  Aiesha Letman and
                  Akhil Mathur and
                  Alan Schelten and
                  Amy Yang and
                  Angela Fan and
                  Anirudh Goyal and
                  Anthony Hartshorn and
                  Aobo Yang and
                  Archi Mitra and
                  Archie Sravankumar and
                  Artem Korenev and
                  Arthur Hinsvark and
                  Arun Rao and
                  Aston Zhang and
                  Aur{\'{e}}lien Rodriguez and
                  Austen Gregerson and
                  Ava Spataru and
                  Baptiste Rozi{\`{e}}re and
                  Bethany Biron and
                  Binh Tang and
                  Bobbie Chern and
                  Charlotte Caucheteux and
                  Chaya Nayak and
                  Chloe Bi and
                  Chris Marra and
                  Chris McConnell and
                  Christian Keller and
                  Christophe Touret and
                  Chunyang Wu and
                  Corinne Wong and
                  Cristian Canton Ferrer and
                  Cyrus Nikolaidis and
                  Damien Allonsius and
                  Daniel Song and
                  Danielle Pintz and
                  Danny Livshits and
                  David Esiobu and
                  Dhruv Choudhary and
                  Dhruv Mahajan and
                  Diego Garcia{-}Olano and
                  Diego Perino and
                  Dieuwke Hupkes and
                  Egor Lakomkin and
                  Ehab AlBadawy and
                  Elina Lobanova and
                  Emily Dinan and
                  Eric Michael Smith and
                  Filip Radenovic and
                  Frank Zhang and
                  Gabriel Synnaeve and
                  Gabrielle Lee and
                  Georgia Lewis Anderson and
                  Graeme Nail and
                  Gr{\'{e}}goire Mialon and
                  Guan Pang and
                  Guillem Cucurell and
                  Hailey Nguyen and
                  Hannah Korevaar and
                  Hu Xu and
                  Hugo Touvron and
                  Iliyan Zarov and
                  Imanol Arrieta Ibarra and
                  Isabel M. Kloumann and
                  Ishan Misra and
                  Ivan Evtimov and
                  Jade Copet and
                  Jaewon Lee and
                  Jan Geffert and
                  Jana Vranes and
                  Jason Park and
                  Jay Mahadeokar and
                  Jeet Shah and
                  Jelmer van der Linde and
                  Jennifer Billock and
                  Jenny Hong and
                  Jenya Lee and
                  Jeremy Fu and
                  Jianfeng Chi and
                  Jianyu Huang and
                  Jiawen Liu and
                  Jie Wang and
                  Jiecao Yu and
                  Joanna Bitton and
                  Joe Spisak and
                  Jongsoo Park and
                  Joseph Rocca and
                  Joshua Johnstun and
                  Joshua Saxe and
                  Junteng Jia and
                  Kalyan Vasuden Alwala and
                  Kartikeya Upasani and
                  Kate Plawiak and
                  Ke Li and
                  Kenneth Heafield and
                  Kevin Stone and
                  et al.},
  title        = {The Llama 3 Herd of Models},
  howpublished   = {arXiv:2407.21783},
  year         = {2024}
}

@inproceedings{ccs/Liu25,
  author       = {Renyang Liu and
                  Wenjie Feng and
                  Tianwei Zhang and
                  Wei Zhou and
                  Xueqi Cheng and
                  See{-}Kiong Ng},
  title        = {Rethinking Machine Unlearning in Image Generation Models},
  booktitle    = {{CCS}},
  year         = {2025}
}

@inproceedings{emnlp/KaramolegkouLZS23,
  author       = {Antonia Karamolegkou and
                  Jiaang Li and
                  Li Zhou and
                  Anders S{\o}gaard},
  title        = {Copyright Violations and Large Language Models},
  booktitle    = {{EMNLP}},
  pages        = {7403--7412},
  year         = {2023}
}

@inproceedings{iclr/LoshchilovH19,
  author       = {Ilya Loshchilov and
                  Frank Hutter},
  title        = {Decoupled Weight Decay Regularization},
  booktitle    = {{ICLR}},
  year         = {2019}
}

@inproceedings{emnlp/LiuZJC24,
  author       = {Yujian Liu and
                  Yang Zhang and
                  Tommi S. Jaakkola and
                  Shiyu Chang},
  title        = {Revisiting Who's Harry Potter: Towards Targeted Unlearning from a
                  Causal Intervention Perspective},
  booktitle    = {{EMNLP}},
  pages        = {8708--8731},
  year         = {2024}
}

@misc{corr/zhang24,
  author       = {Ruiqi Zhang and
                  Licong Lin and
                  Yu Bai and
                  Song Mei},
  title        = {Negative Preference Optimization: From Catastrophic Collapse to Effective
                  Unlearning},
  howpublished   = {arXiv:2404.05868},
  year         = {2024}
}

@article{harding2019,
  title={Understanding the scope and impact of the california consumer privacy act of 2018},
  author={Harding, Elizabeth Liz and Vanto, Jarno J and Clark, Reece and Hannah Ji, L and Ainsworth, Sara C},
  journal={Journal of Data Protection \& Privacy},
  volume={2},
  number={3},
  pages={234--253},
  year={2019}
}

@article{mantelero2013,
  title={The EU Proposal for a General Data Protection Regulation and the roots of the ‘right to be forgotten’},
  author={Mantelero, Alessandro},
  journal={Computer Law \& Security Review},
  volume={29},
  number={3},
  pages={229--235},
  year={2013},
}

@inproceedings{iclr/gao25,
  title={On Large Language Model Continual Unlearning},
  author={Gao, Chongyang and Wang, Lixu and Ding, Kaize and Weng, Chenkai and Wang, Xiao and Zhu, Qi},
  year={2025},
  booktitle={ICLR},
}

@article{natmi/LiuYJCBHYLXLVBKL25,
  author       = {Sijia Liu and
                  Yuanshun Yao and
                  Jinghan Jia and
                  Stephen Casper and
                  Nathalie Baracaldo and
                  Peter Hase and
                  Yuguang Yao and
                  Chris Yuhao Liu and
                  Xiaojun Xu and
                  Hang Li and
                  Kush R. Varshney and
                  Mohit Bansal and
                  Sanmi Koyejo and
                  Yang Liu},
  title        = {Rethinking machine unlearning for large language models},
  journal      = {Nat. Mac. Intell.},
  volume       = {7},
  number       = {2},
  pages        = {181--194},
  year         = {2025}
}

@misc{corr/Barez25,
  author       = {Fazl Barez and
                  Tingchen Fu and
                  Ameya Prabhu and
                  Stephen Casper and
                  Amartya Sanyal and
                  Adel Bibi and
                  Aidan O'Gara and
                  Robert Kirk and
                  Ben Bucknall and
                  Tim Fist and
                  Luke Ong and
                  Philip Torr and
                  Kwok{-}Yan Lam and
                  Robert Trager and
                  David Krueger and
                  S{\"{o}}ren Mindermann and
                  Jos{\'{e}} Hern{\'{a}}ndez{-}Orallo and
                  Mor Geva and
                  Yarin Gal},
  title        = {Open Problems in Machine Unlearning for {AI} Safety},
  howpublished   = {arXiv:2501.04952},
  year         = {2025}
}

@misc{corr/Young24,
  author       = {Alex Young and
                  Bei Chen and
                  Chao Li and
                  Chengen Huang and
                  Ge Zhang and
                  Guanwei Zhang and
                  Heng Li and
                  Jiangcheng Zhu and
                  Jianqun Chen and
                  Jing Chang and
                  Kaidong Yu and
                  Peng Liu and
                  Qiang Liu and
                  Shawn Yue and
                  Senbin Yang and
                  Shiming Yang and
                  Tao Yu and
                  Wen Xie and
                  Wenhao Huang and
                  Xiaohui Hu and
                  Xiaoyi Ren and
                  Xinyao Niu and
                  Pengcheng Nie and
                  Yuchi Xu and
                  Yudong Liu and
                  Yue Wang and
                  Yuxuan Cai and
                  Zhenyu Gu and
                  Zhiyuan Liu and
                  Zonghong Dai},
  title        = {Yi: Open Foundation Models by 01.AI},
  howpublished   = {arXiv:2403.04652},
  year         = {2024}
}

@inproceedings{iclr/ZhengCQ025,
  author       = {Junhao Zheng and
                  Xidi Cai and
                  Shengjie Qiu and
                  Qianli Ma},
  title        = {Spurious Forgetting in Continual Learning of Language Models},
  booktitle    = {{ICLR}},
  year         = {2025}
}

@misc{corr/cobbe21,
  author       = {Karl Cobbe and
                  Vineet Kosaraju and
                  Mohammad Bavarian and
                  Mark Chen and
                  Heewoo Jun and
                  Lukasz Kaiser and
                  Matthias Plappert and
                  Jerry Tworek and
                  Jacob Hilton and
                  Reiichiro Nakano and
                  Christopher Hesse and
                  John Schulman},
  title        = {Training Verifiers to Solve Math Word Problems},
  howpublished   = {arXiv:2110.14168},
  year         = {2021}
}

@inproceedings{iclr/HendrycksBBZMSS21,
  author       = {Dan Hendrycks and
                  Collin Burns and
                  Steven Basart and
                  Andy Zou and
                  Mantas Mazeika and
                  Dawn Song and
                  Jacob Steinhardt},
  title        = {Measuring Massive Multitask Language Understanding},
  booktitle    = {{ICLR}},
  year         = {2021}
}

@inproceedings{naacl/TalmorHLB19,
  author       = {Alon Talmor and
                  Jonathan Herzig and
                  Nicholas Lourie and
                  Jonathan Berant},
  title        = {CommonsenseQA: {A} Question Answering Challenge Targeting Commonsense
                  Knowledge},
  booktitle    = {{NAACL}},
  pages        = {4149--4158},
  year         = {2019}
}

@inproceedings{nips/lu22learn,
  title={Learn to explain: Multimodal reasoning via thought chains for science question answering},
  author={Lu, Pan and Mishra, Swaroop and Xia, Tanglin and Qiu, Liang and Chang, Kai-Wei and Zhu, Song-Chun and Tafjord, Oyvind and Clark, Peter and Kalyan, Ashwin},
  booktitle = {{NeurIPS}},
  pages={2507--2521},
  year={2022}
}

@inproceedings{ijcai/RozemberczkiWBY22,
  author       = {Benedek Rozemberczki and
                  Lauren Watson and
                  P{\'{e}}ter Bayer and
                  Hao{-}Tsung Yang and
                  Oliver Kiss and
                  Sebastian Nilsson and
                  Rik Sarkar},
  title        = {The Shapley Value in Machine Learning},
  booktitle    = {{IJCAI}},
  pages        = {5572--5579},
  year         = {2022}
}

@inproceedings{nips/ZhaoKBTT24,
  author       = {Kairan Zhao and
                  Meghdad Kurmanji and
                  George{-}Octavian Barbulescu and
                  Eleni Triantafillou and
                  Peter Triantafillou},
  title        = {What makes unlearning hard and what to do about it},
  booktitle    = {{NeurIPS}},
  year         = {2024}
}

@inproceedings{iclr/AnconaCO018,
  author       = {Marco Ancona and
                  Enea Ceolini and
                  Cengiz {\"{O}}ztireli and
                  Markus Gross},
  title        = {Towards better understanding of gradient-based attribution methods
                  for Deep Neural Networks},
  booktitle    = {{ICLR}},
  year         = {2018}
}

@inproceedings{emnlp/GaoYC21,
  author       = {Tianyu Gao and
                  Xingcheng Yao and
                  Danqi Chen},
  title        = {SimCSE: Simple Contrastive Learning of Sentence Embeddings},
  booktitle    = {{EMNLP}},
  pages        = {6894--6910},
  year         = {2021}
}

@inproceedings{emnlp/xu25,
  author={Xu, Xiaoyu and Du, Minxin and Ye, Qingqing and Hu, Haibo},
  title={OBLIVIATE: Robust and Practical Machine Unlearning for Large Language Models},
  booktitle    = {{EMNLP}},
  year         = {2025}
}

@inproceedings{iclr/YuanPDC0L25,
  author       = {Xiaojian Yuan and
                  Tianyu Pang and
                  Chao Du and
                  Kejiang Chen and
                  Weiming Zhang and
                  Min Lin},
  title        = {A Closer Look at Machine Unlearning for Large Language Models},
  booktitle    = {{ICLR}},
  year         = {2025}
}

@inproceedings{iclr/ZhangWLWTL00W25,
  author       = {Zhiwei Zhang and
                  Fali Wang and
                  Xiaomin Li and
                  Zongyu Wu and
                  Xianfeng Tang and
                  Hui Liu and
                  Qi He and
                  Wenpeng Yin and
                  Suhang Wang},
  title        = {Catastrophic Failure of {LLM} Unlearning via Quantization},
  booktitle    = {{ICLR}},
  year         = {2025}
}

@inproceedings{icml/xu2025,
  author={Xu, Xiaoyu and Yue, Xiang and Liu, Yang and Ye, Qingqing and Zheng, Huadi and Hu, Peizhao and Du, Minxin and Hu, Haibo},
  title={Unlearning Isn't Deletion: Investigating Reversibility of Machine Unlearning in LLMs},
  booktitle    = {{ICML}},
  year         = {2026}
}

@misc{corr/goel2022towards,
  title={Towards adversarial evaluations for inexact machine unlearning},
  author={Goel, Shashwat and Prabhu, Ameya and Sanyal, Amartya and Lim, Ser-Nam and Torr, Philip and Kumaraguru, Ponnurangam},
  howpublished  = {arXiv:2201.06640},
  year={2022}
}

@inproceedings{icml/wuerkaixi2025adaptive,
  title={Adaptive localization of knowledge negation for continual llm unlearning},
  author={Wuerkaixi, Abudukelimu and Wang, Qizhou and Cui, Sen and Xu, Wutong and Han, Bo and Niu, Gang and Sugiyama, Masashi and Zhang, Changshui},
  booktitle={{ICML}},
  year={2025}
}

@misc{corr/Li2505,
  author       = {Zexi Li and
                  Xiangzhu Wang and
                  William F. Shen and
                  Meghdad Kurmanji and
                  Xinchi Qiu and
                  Dongqi Cai and
                  Chao Wu and
                  Nicholas D. Lane},
  title        = {Editing as Unlearning: Are Knowledge Editing Methods Strong Baselines
                  for Large Language Model Unlearning?},
  howpublished  = {arXiv:2505.19855},
  year         = {2025}
}

@inproceedings{usenix/Ye25,
  author       = {Dayong Ye and
                  Tainqing Zhu and
                  Jiayang Li and
                  Kun Gao and
                  Bo Liu and
                  Leo Yu Zhang and
                  Wanlei Zhou and
                  Yang Zhang},
  title        = {Data Duplication: {A} Novel Multi-Purpose Attack Paradigm in Machine
                  Unlearning},
  booktitle={{USENIX Security}},
  year         = {2025}
}

@inproceedings{usenix/Song25,
  author       = {Minkyoo Song and Hanna Kim and Jaehan Kim and Seungwon Shin and Sooel Son},
  title        = {Refusal Is Not an Option: Unlearning Safety Alignment of Large Language Models},
  booktitle=   {{USENIX Security}},
  year         = {2025}
}

@inproceedings{sp/Xia25,
  author       = {Xiaoyu Xia and
                  Ziqi Wang and
                  Ruoxi Sun and
                  Bowen Liu and
                  Ibrahim Khalil and
                  Minhui Xue},
  title        = {Edge Unlearning is Not "on Edge"! an Adaptive Exact Unlearning System
                  on Resource-Constrained Devices},
  booktitle    = {{S\&P}},
  pages        = {2546--2563},
  year         = {2025}
}

@inproceedings{usenix/long25,
  author       = {Cheng{-}Long Wang and
                  Qi Li and
                  Zihang Xiang and
                  Yinzhi Cao and
                  Di Wang},
  title        = {Towards Lifecycle Unlearning Commitment Management: Measuring Sample-level
                  Unlearning Completeness},
  booktitle={{USENIX Security}},
  year         = {2025}
}

@inproceedings{sp/HuWDX24,
  author       = {Hongsheng Hu and
                  Shuo Wang and
                  Tian Dong and
                  Minhui Xue},
  title        = {Learn What You Want to Unlearn: Unlearning Inversion Attacks against
                  Machine Unlearning},
  booktitle    = {{S\&P}},
  pages        = {3257--3275},
  year         = {2024}
}

@inproceedings{nips/JinCWHYL00024,
  author       = {Zhuoran Jin and
                  Pengfei Cao and
                  Chenhao Wang and
                  Zhitao He and
                  Hongbang Yuan and
                  Jiachun Li and
                  Yubo Chen and
                  Kang Liu and
                  Jun Zhao},
  title        = {{RWKU:} Benchmarking Real-World Knowledge Unlearning for Large Language
                  Models},
  booktitle    = {{NeurIPS}},
  year         = {2024}
}

@inproceedings{acl/LoBC24,
  author       = {Michelle Lo and
                  Fazl Barez and
                  Shay B. Cohen},
  title        = {Large Language Models Relearn Removed Concepts},
  booktitle    = {Findings of {ACL}},
  pages        = {8306--8323},
  year         = {2024}
}

@inproceedings{sp/CaoY15,
  author       = {Yinzhi Cao and
                  Junfeng Yang},
  title        = {Towards Making Systems Forget with Machine Unlearning},
  booktitle    = {{S\&P}},
  pages        = {463--480},
  year         = {2015}
}

@inproceedings{sigir/ShiX0Z00025,
  author       = {Teng Shi and
                  Jun Xu and
                  Xiao Zhang and
                  Xiaoxue Zang and
                  Kai Zheng and
                  Yang Song and
                  Han Li},
  title        = {Retrieval Augmented Generation with Collaborative Filtering for Personalized
                  Text Generation},
  booktitle    = {{SIGIR}},
  pages        = {1294--1304},
  year         = {2025}
}

@inproceedings{nips/LewisPPPKGKLYR020,
  author       = {Patrick Lewis and
                  Ethan Perez and
                  Aleksandra Piktus and
                  Fabio Petroni and
                  Vladimir Karpukhin and
                  Naman Goyal and
                  Heinrich K{\"{u}}ttler and
                  Mike Lewis and
                  Wen{-}tau Yih and
                  Tim Rockt{\"{a}}schel and
                  Sebastian Riedel and
                  Douwe Kiela},
  title        = {Retrieval-Augmented Generation for Knowledge-Intensive {NLP} Tasks},
  booktitle    = {{NeurIPS}},
  year         = {2020}
}

@misc{corr/Thakral25,
  author       = {Kartik Thakral and
                  Tamar Glaser and
                  Tal Hassner and
                  Mayank Vatsa and
                  Richa Singh},
  title        = {Continual Unlearning for Foundational Text-to-Image Models without
                  Generalization Erosion},
  howpublished  = {arXiv:2503.13769},
  year         = {2025}
}

@inproceedings{icml/lee2025an,
    title={An Empirical Exploration of Continual Unlearning for Image Generation},
    author={Justin Lee and Zheda Mai and Chongyu Fan and Wei-Lun Chao},
    booktitle={{ICML} 2025 Workshop on Machine Unlearning for Generative AI},
    year={2025},
}

@inproceedings{iclr/0001FWS25,
  author       = {Shengyuan Hu and
                  Yiwei Fu and
                  Steven Z. Wu and
                  Virginia Smith},
  title        = {Unlearning or Obfuscating? Jogging the Memory of Unlearned LLMs via
                  Benign Relearning},
  booktitle    = {{ICLR}},
  year         = {2025}
}

@inproceedings{iclr/shi24,
  title        = {{MUSE:} Machine Unlearning Six-Way Evaluation for Language Models},
  author       = {Weijia Shi and
                  Jaechan Lee and
                  Yangsibo Huang and
                  Sadhika Malladi and
                  Jieyu Zhao and
                  Ari Holtzman and
                  Daogao Liu and
                  Luke Zettlemoyer and
                  Noah A. Smith and
                  Chiyuan Zhang},
  booktitle={ICLR},
  year={2025},
  
}

@inproceedings{CVPR/zhao2024,
  title={Continual forgetting for pre-trained vision models},
  author={Zhao, Hongbo and Ni, Bolin and Fan, Junsong and Wang, Yuxi and Chen, Yuntao and Meng, Gaofeng and Zhang, Zhaoxiang},
  booktitle={CVPR},
  pages={28631--28642},
  year={2024}
}

@inproceedings{emnlp/gur2025,
  title={Precise in-parameter concept erasure in large language models},
  author={Gur-Arieh, Yoav and Suslik, Clara Haya and Hong, Yihuai and Barez, Fazl and Geva, Mor},
  booktitle={{EMNLP}},
  pages={18986–19006},
  year={2025}
}

@misc{corr/weiGu24,
  author       = {Jiawei Gu and
                  Xuhui Jiang and
                  Zhichao Shi and
                  Hexiang Tan and
                  Xuehao Zhai and
                  Chengjin Xu and
                  Wei Li and
                  Yinghan Shen and
                  Shengjie Ma and
                  Honghao Liu and
                  Yuanzhuo Wang and
                  Jian Guo},
  title        = {A Survey on LLM-as-a-Judge},
  howpublished  = {arXiv:2411.15594},
  year         = {2024}
}

@inproceedings{iclr/FangJWMSW0C25,
  author       = {Junfeng Fang and
                  Houcheng Jiang and
                  Kun Wang and
                  Yunshan Ma and
                  Jie Shi and
                  Xiang Wang and
                  Xiangnan He and
                  Tat{-}Seng Chua},
  title        = {AlphaEdit: Null-Space Constrained Knowledge Editing for Language Models},
  booktitle    = {{ICLR}},
  year         = {2025}
}

@misc{2023opencompass,
    title={OpenCompass: A Universal Evaluation Platform for Foundation Models},
    author={OpenCompass Contributors},
    howpublished = {\url{https://github.com/open-compass/opencompass}},
    year={2023}
}

@misc{corr/Hurst2024GPT4oSC,
  title={GPT-4o System Card},
  author={OpenAI},
  howpublished  = {arXiv:2410.21276},
  year={2024}
}

@inproceedings{acl/Bae26,
  title={CURaTE: Continual Unlearning in Real Time with Ensured Preservation of LLM Knowledge},
  author={Seyun Bae and Seokhan Lee and Eunho Yang},
  booktitle    = {Findings of {ACL}},
  year         = {2026}
}

@misc{corr/Yang25,
  author={An Yang and Anfeng Li and Baosong Yang and Beichen Zhang and Binyuan Hui and Bo Zheng and Bowen Yu and Chang Gao and Chengen Huang and Chenxu Lv and Chujie Zheng and Dayiheng Liu and Fan Zhou and Fei Huang and Feng Hu and Hao Ge and Haoran Wei and Huan Lin and Jialong Tang and Jian Yang and Jianhong Tu and Jianwei Zhang and Jianxin Yang and Jiaxin Yang and Jingren Zhou and Jingren Zhou and Junyan Lin and Kai Dang and Keqin Bao and Ke‐Pei Yang and Le Yu and Li-Chun Deng and Mei Li and Min Xue and Mingze Li and Pei Zhang and Peng Wang and Qin Zhu and Rui Men and Ruize Gao and Shi-Qiang Liu and Shuang Luo and Tianhao Li and Tianyi Tang and Wenbiao Yin and Xingzhang Ren and Xinyu Wang and Xinyu Zhang and Xuancheng Ren and Yang Fan and Yang Su and Yi-Chao Zhang and Yinger Zhang and Yu Wan and Yuqiong Liu and Zekun Wang and Zeyu Cui and Zhenru Zhang and Zhipeng Zhou and Zihan Qiu},
  title={Qwen3 Technical Report},
  howpublished   = {arXiv:2505.09388},
  year         = {2025}
}
\clearpage 
\appendix  

\section{Limitations}\label{sec:limitations}
Our evaluation of \fit spans models up to 14B parameters, representing a practical and widely adopted regime for open-source LLM research. 
However, validating the framework's scalability on 70B+ parameter models remains an important direction for future work, currently limited by computational constraints. 
Furthermore, our continual unlearning streams and evaluation checkpoints are predefined to ensure controlled, reproducible baseline comparisons. 
In practice, real-world deletion requests may exhibit more complex temporal dependencies, varying arrival rates, and broader semantic distributions. 
Future research should extend \fit to longer interaction horizons and investigate its behavior under uncurated, highly dynamic request streams.

\section{Broader Impacts}\label{Broader impacts}
This work advances the reliability of LLM unlearning under continual, high-volume deletion requests. 
By mitigating catastrophic forgetting, we offer technical solutions to better support privacy protection, copyright compliance, and safety alignment (\eg, facilitating adherence to GDPR's ``Right to be Forgotten''). 
Furthermore, by formalizing the risks of post-unlearning recovery, \fit and \textbf{PCH} equip the community with rigorous tools to systematically evaluate the true efficacy of unlearning.

Despite these positive impacts, approximate machine unlearning carries inherent risks. 
A primary concern is the potential for a false sense of security: over-reliance on approximate unlearning as definitive proof of data deletion could lead to severe privacy or regulatory failures if residual knowledge remains extractable. 
Additionally, malicious actors might exploit unlearning mechanisms to deliberately obscure model provenance or erase accountability-relevant information. 
We also caution that the synthetic proxy retain models used for evaluation may not perfectly mirror true retraining behavior, which could lead to miscalibrated deployment decisions. 
Finally, while the harmful-content samples in \textbf{PCH} are strictly synthetic and non-operational, deploying unlearning in real-world production environments necessitates rigorous auditing, continual monitoring, and comprehensive legal review prior to processing authentic user, copyrighted, or safety-critical data.

\section{Declaration of LLM Usage}\label{LLM_USAGE}
We utilized GPT-4o within our dataset construction and validation pipelines. 
Specifically, GPT-4o was employed to generate the synthetic \textbf{PCH} dataset according to the structured protocols detailed in Section~\ref{dataset}. 
Furthermore, we used a GPT-4o-as-a-judge evaluation to assess sensitivity scores and verify that our two-stage filtering mechanism successfully prunes redundant context while preserving sensitive target tokens (\eg, personal identifiers, harmful terms, and copyrighted expressions). 
All model-generated data and automated evaluations were rigorously inspected and manually verified by the authors, who assume full responsibility for the contents of this paper.

\section{More Details on \fit}
\label{fit_details}
This section presents the mathematical foundations of \fit's three core components: embedding-based redundancy filtering, importance-guided algorithm selection, and targeted layer attribution.

\subsection{Embedding-Based Redundancy Filtering}\label{filtering_proof}

Consider an incoming request at round \(t{+}1\) containing a chunk \(\{x_i\} \subset \mathcal{D}_f^{(t+1)}\) and a historical set \(\mathsf{D}_f^{(1:t)}\) containing a chunk \(\{x_j\}\). 
Let \(g_i=\nabla_\theta L(x_i)\) and \(g_j=\nabla_\theta L(x_j)\) denote their respective gradients.
When two chunks are semantically redundant, their embeddings often exhibit high cosine similarity, which we use as a practical proxy for potential gradient overlap:
\[
\cos(\mathsf{e}(x_i),\mathsf{e}(x_j))\approx 1
\quad \Rightarrow \quad
g_i^\top g_j > 0 \ \text{in expectation}.
\]
In the idealized case where redundant chunks induce nearly aligned gradients, the accumulated update behaves like
$
g_{\mathrm{agg}} \approx n g(x),
$
and the empirical Fisher can be approximated as
$
I(X)=\sum_{i=1}^{n} g_i g_i^\top \approx n\,g(x)g(x)^\top .
$
This approximation suggests an amplified dominant direction in the local curvature, providing an intuition for why repeated redundant requests may steepen the local loss landscape and increase the risk of catastrophic forgetting.

While filtering based on embedding similarity mitigates this rank-1 amplification, similarity alone is an insufficient proxy for redundancy. 
Distinct sequences may share lexical structures (\eg, ``My name is Alice'' vs.\ ``My name is Bob'') yet contain unique sensitive tokens, such as personally identifiable information or harmful terms, that necessitate removal. To prevent the erroneous preservation of sensitive data, we introduce a loss-difference test:
$
\Delta L(x) = \left| L_{\text{with}} - L_{\text{without}} \right|.
$
Inspired by influence functions, it serves as a practical surrogate for estimating a sample's semantic contribution to the model. 
Samples with large \(\Delta L\) are likely to induce substantial changes in model predictions and are retained for unlearning, regardless of their embedding similarity to history.
  
\textbf{Discussion.}
Existing retrieval-augmented generation (RAG) systems typically employ similarity-based filtering to remove redundant context~\cite{nips/LewisPPPKGKLYR020,sigir/ShiX0Z00025}. 
However, these methods fail to account for the semantic role of sensitive tokens. 
By incorporating the \(\Delta L\) test, our approach distinguishes between structural redundancy and sensitive information.
We further adopt rare-token filtering~\cite{uss/CarliniTWJHLRBS21} to protect segments containing tokens uncommon in the pre-training corpus, as such tokens are often highly memorized and privacy-critical. 
While effective, rare-token statistics for proprietary corpora are often unavailable; future work will explore model-internal proxies for rarity estimation.

\begin{figure}[!t]
  \centering
  \includegraphics[width=0.8\linewidth]{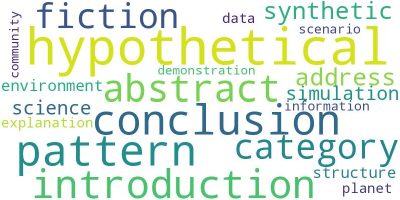}
  \caption{Word cloud of the semantic content in the filtered forget data. 
  Each word’s font size is proportional to its normalized frequency (larger font $\Rightarrow$ higher frequency) after filtering. 
  The prevalence of neutral terms—\textit{hypothetical}, \textit{pattern}, \textit{conclusion}—indicates that the pipeline removes
    irrelevant text while preserving potentially sensitive information.}
  \label{fig:wordcloud-semantic}
\end{figure}

\begin{algorithm}[!t]
\caption{Similar Embedding Filtering}
\label{alg:embedding-filtering}
\begin{algorithmic}[1]
\REQUIRE Unlearning request $\mathcal{D}_f^{(t+1)}$, historical forget set $\mathsf{D}_f^{(1:t)}$, embedding model $\mathsf{e}(\cdot)$, and target LLM~$\mathcal{M}$
\ENSURE Filtered request $\mathcal{D}_f^{(t+1)*}$, new history $\mathsf{D}_f^{(1:t+1)}$
\PP Chunk size $c$, similarity and loss thresholds $\tau$ and $\epsilon$ 
\STATE Initialize $\mathcal{D}_f^{(t+1)*} \gets \emptyset$
\STATE \label{line:chunk-split} Split $\mathcal{D}_f^{(t+1)}$ into chunks $x$ of size $c$
\STATE \label{line:mem-embed} Fetch all memory embeddings $\mathsf{e}(m)$, $\forall m \in \mathsf{D}_f^{(1:t)}$ 
\FOR{each $x$} \label{line:loop-start}
    \STATE \label{line:sim-threshold} Compute embedding $\mathsf{e}(x)$
    \STATE Record $s^* = \max\limits_{m} \cos (\mathsf{e}(x), \mathsf{e}(m))$, with $m \in \mathsf{D}_f^{(1:t)}$
    \IF{$s^* < \tau$} 
        \STATE \label{line:sim-pass} Append $x$ to $\mathcal{D}_f^{(t+1)*}$
    \ELSE
        \STATE \label{line:loss-check-start} Compute $L_{\mathrm{with}} = \mathrm{CE}(\mathcal{D}_f^{(t+1)} , \mathcal{M})$
        \STATE Compute $L_{\mathrm{without}} = \mathrm{CE}(\mathcal{D}_f^{(t+1)} \setminus x, \mathcal{M})$
        \IF{$|L_{\mathrm{with}} - L_{\mathrm{without}}| > \epsilon$} \label{line:loss-pass}
            \STATE Append $x$ to $\mathcal{D}_f^{(t+1)*}$ \label{line:loss-end}
        \ENDIF \label{line:loss-check-end}
    \ENDIF
\ENDFOR \label{line:loop-end}
\STATE \textbf{return} $\mathcal{D}_f^{(t+1)*}, \mathsf{D}_f^{(1:t+1)} =\ \mathsf{D}_f^{(1:t)} \cup \mathcal{D}_f^{(t+1)*}$
\end{algorithmic}
\end{algorithm}
\begin{wraptable}{r}{0.52\textwidth}
    \centering
    \vspace{-10pt}
    \caption{Sensitivity of data filtering under different random seeds}
    \label{tab:sensitivity}
    \resizebox{0.50\textwidth}{!}{
    \begin{tabular}{lcc}
    \toprule
    \textbf{Model} & \textbf{Seeds} & \textbf{Sensitivity Avg.} \\
    \midrule
    Llama-2-7b-chat-hf  & 20, 30, 40, 50 & 2.00 \\
    Llama-3-8B          & 20, 30, 40, 50 & 1.50 \\
    Llama-3-8B-Instruct & 20, 30, 40, 50 & 1.50 \\
    Yi-6B               & 20, 30, 40, 50 & 1.75 \\
    \bottomrule
    \end{tabular}
    }
    \vspace{-5pt}
\end{wraptable}
\textbf{Data Filtering Analysis.} 
\label{filtering}
To validate the filtering module, we incorporated stress-test cases into our dataset, such as individuals sharing a name (\eg, \textit{Liam Hawthorne}) but differing in attributes. 
This ensures the module removes genuinely redundant context while preserving distinct sensitive identifiers.
A sensitivity evaluation using GPT-4o as a judge~\cite{corr/weiGu24} yielded average sensitivity scores consistently close to 1.0 (on a 5-point scale) across multiple seeds (Table~\ref{tab:sensitivity}), confirming the safety of the filtered sets. 
Furthermore, a frequency analysis of the filtered text (Figure~\ref{fig:wordcloud-semantic}) reveals that removed terms are predominantly neutral (\eg, \textit{description}, \textit{context}).

\subsection{Importance-Guided Algorithm Selection}\label{impor}

We now provide a formal justification for using the input \texttt{IMP} score as a proxy for local request sensitivity.
Our goal here is not to claim that \texttt{IMP} exactly recovers global memorization or full parameter-space unlearning sensitivity.
Instead, we show that \texttt{IMP} controls the first-order local variation of the training loss with respect to perturbations in the request representation, up to a second-order smoothness remainder, which directly motivates importance-guided algorithm selection.

Let \(x^* = \mathcal{D}_f^{(t+1)*}\) denote the filtered forget request, and let
\[
z = E(x^*)
\]
be its input embedding representation.
We write the loss as \(L(z)\), and define the importance score as
\[
\texttt{IMP}(x^*) = \|g(x^*)\|_2,
\qquad
g(x^*) = \nabla_z L(z)\big|_{z = E(x^*)}.
\]

\paragraph{Proposition 1.}
Assume that \(L:\mathbb{R}^d \to \mathbb{R}\) is differentiable and \(\beta\)-smooth in a neighborhood of \(z\), i.e.,
\[
\|\nabla L(z_1)-\nabla L(z_2)\|_2 \le \beta \|z_1-z_2\|_2
\]
for all \(z_1,z_2\) in that neighborhood.
Then for any perturbation \(\delta \in \mathbb{R}^d\) such that \(z+t\delta\) remains in the same neighborhood for all \(t\in[0,1]\),
\[
|L(z+\delta)-L(z)|
\le
\|\nabla L(z)\|_2 \,\|\delta\|_2
+
\frac{\beta}{2}\|\delta\|_2^2.
\]
Equivalently,
\[
|L(E(x^*)+\delta)-L(E(x^*))|
\le
\texttt{IMP}(x^*)\,\|\delta\|_2
+
\frac{\beta}{2}\|\delta\|_2^2.
\]

\paragraph{Proof.}
Since \(L\) is differentiable, we can write
\[
L(z+\delta)-L(z)
=
\int_0^1 \nabla L(z+t\delta)^\top \delta \, dt.
\]
Adding and subtracting \(\nabla L(z)\), we obtain
\[
L(z+\delta)-L(z)
=
\nabla L(z)^\top \delta
+
\int_0^1 \big(\nabla L(z+t\delta)-\nabla L(z)\big)^\top \delta \, dt.
\]
Therefore,
\[
\left|L(z+\delta)-L(z)-\nabla L(z)^\top \delta\right|
\le
\int_0^1
\left\|\nabla L(z+t\delta)-\nabla L(z)\right\|_2 \,\|\delta\|_2 \, dt.
\]
By \(\beta\)-smoothness,
\[
\left\|\nabla L(z+t\delta)-\nabla L(z)\right\|_2
\le
\beta \|t\delta\|_2
=
\beta t \|\delta\|_2,
\]
which implies
\[
\left|L(z+\delta)-L(z)-\nabla L(z)^\top \delta\right|
\le
\int_0^1 \beta t \|\delta\|_2^2 \, dt
=
\frac{\beta}{2}\|\delta\|_2^2.
\]
Hence,
\[
|L(z+\delta)-L(z)|
\le
|\nabla L(z)^\top \delta|
+
\frac{\beta}{2}\|\delta\|_2^2.
\]
By the Cauchy--Schwarz inequality,
\[
|\nabla L(z)^\top \delta|
\le
\|\nabla L(z)\|_2\,\|\delta\|_2.
\]
Thus,
\[
|L(z+\delta)-L(z)|
\le
\|\nabla L(z)\|_2\,\|\delta\|_2
+
\frac{\beta}{2}\|\delta\|_2^2.
\]
Substituting \(\nabla L(z)=g(x^*)\) and \(z=E(x^*)\) gives
\[
|L(E(x^*)+\delta)-L(E(x^*))|
\le
\texttt{IMP}(x^*)\,\|\delta\|_2
+
\frac{\beta}{2}\|\delta\|_2^2.
\]
This completes the proof. Proposition~1 shows that \(\texttt{IMP}(x^*)\) determines the magnitude of the first-order local loss variation in embedding space, up to a second-order smoothness remainder.
Hence, larger \texttt{IMP} scores correspond to steeper local loss geometry around the request representation.
This suggests that, in continual unlearning, higher-\texttt{IMP} requests may induce stronger optimization responses under repeated updates, which motivates routing high-\texttt{IMP} requests to more conservative objectives and low-\texttt{IMP} requests to more aggressive ones.

\textbf{Discussion.}
Prior work has shown that deletion efficacy is related to memorization strength~\cite{nips/ZhaoKBTT24}.
Our analysis does not claim that \texttt{IMP} is an exact memorization metric.
Rather, it provides a rigorous justification that \texttt{IMP} is a lightweight, gradient-based indicator of \emph{local request sensitivity}, which is the quantity most directly relevant to stable routing in continual unlearning.

\subsection{Targeted Layer Attribution}
\label{layer}

\begin{algorithm}[!t]
\caption{Targeted Layer Attribution}
\label{alg:layer-attribution}
\begin{algorithmic}[1]
\REQUIRE \label{line:attr-input-start}  filtered forget set $\mathcal{D}_f^{(t+1)*}$, number of layers $L$, and~$ \mathcal{M}$
\ENSURE \label{line:attr-output-start} Targeted layer indices $\mathcal{S}_{\text{top-}K\%}$ \label{line:attr-output-end}
\STATE \label{line:loss-orig} Compute original loss: $L_{\text{orig}} \gets \mathrm{CE}(\mathcal{D}_f^{(t+1)*}, \mathcal{M})$

\FOR{\label{line:attr-loop-start} each layer index $\ell = 1$ to $L$}
    \STATE \label{line:mask-start} Temporarily mask parameters of layer $\ell$ in $\mathcal{M}$ (\eg, set weights to zero)
    \STATE \label{line:loss-mask} Compute masked loss: 
    
    $L_{\text{mask}}^{(\ell)} \gets \mathrm{CE}(\mathcal{D}_f^{(t+1)*}, \mathcal{M_\text{mask}})$
    \STATE \label{line:restore} Restore parameters of layer $\ell$ in $\mathcal{M}$
    \STATE \label{line:score} Compute attribution score: $s_\ell \gets \left| L_{\text{mask}}^{(\ell)} - L_{\text{orig}} \right|$
\ENDFOR

\STATE \label{line:rank} Rank layers by $s_\ell$ in descending order 
\STATE \label{line:select-topk} $\mathcal{S}_{\text{top-}K\%} \gets$ indices of the top 
$\left\lceil K L / 100 \right\rceil$ layers with highest $s_\ell$
\STATE \textbf{return} \label{line:attr-return} $\mathcal{S}_{\text{top-}K\%}$
\end{algorithmic}
\end{algorithm}

Continual unlearning presents a trade-off: updating all layers is computationally expensive and can harm utility, while overly sparse updates may leave knowledge recoverable.
We follow the principle of \emph{minimal intervention}, modifying only the layers most relevant to each unlearning request.
However, a fixed layer count may not scale well with model size, since forget-related information can be distributed across more layers in larger models.
We therefore adopt a ratio-based rule: for a model with \(L\) layers, FIT selects the top-\(K\%\) most relevant layers among all layers, i.e.,
$
|\mathcal{S}_{\text{top-}K\%}|=\left\lceil KL/100 \right\rceil .
$
Thus, \(K\%\) specifies an update ratio rather than an absolute layer count, allowing the number of updated layers to scale with model size.

To identify \(\mathcal{S}_{\text{top-}K\%}\), we estimate the contribution of layer \(\ell\) by masking it and computing the loss deviation:
$
s_\ell = \left| L_{\text{mask}}^{(\ell)} - L_{\text{orig}} \right|.
$
This leave-one-out score provides a lightweight Shapley-style approximation, rather than an exact Shapley value over all layer coalitions: 
\[
\phi_\ell =
\mathbb{E}_{S\subseteq\mathcal{L}\setminus\{\ell\}}
\left[L(S\cup\{\ell\}) - L(S)\right],
\]
capturing the marginal effect of each layer on the current unlearning request.
FIT then ranks layers by \(s_\ell\) and updates the MLP blocks and attention modules in the top-\(K\%\) layers, while freezing all other parameters.
This provides a tractable surrogate for sparse intervention, focusing updates on forget-relevant components while limiting computational overhead and parameter drift.
Empirically, Figure~\ref{fig:layer_histogram} shows that attribution scores \(\{s_\ell\}\) consistently concentrate on compact layer subsets.

\paragraph{Discussion.}
Recent model-editing methods typically identify neuron- or parameter-level regions associated with specific knowledge and edit them directly.
For example, AlphaEdit~\cite{iclr/FangJWMSW0C25} restricts updates to null-space directions, while PISCES~\cite{emnlp/gur2025} removes concepts through feature-level masking.
These methods rely on narrowly localized parameter subsets and implicitly assume that the relevant parameters are spatially concentrated.
This assumption is reasonable for inserting or modifying isolated facts, where preserving existing knowledge is the main priority and post-unlearning recovery is not central.
However, it may not hold for unlearning, where targeted knowledge can be distributed across multiple layers and effective removal requires sufficient coverage of all relevant regions.

We discuss these methods as related localization strategies, but their assumptions become limiting in continual unlearning.
PISCES further shows that AlphaEdit is vulnerable to post-unlearning recovery.
Although PISCES improves robustness, it still exhibits substantial recoverability under relearning attacks, since parameter-level masks cannot adapt to shifting influence patterns.
This highlights the limitations of fine-grained model-editing localization in continual unlearning.
In contrast, our attribution-guided framework operates at the layer level and estimates relevance dynamically at each unlearning step.
This avoids the brittleness of neuron- or parameter-level edits and provides a stable balance between robustness and efficiency, while remaining compatible with localization signals.

We also need to determine which components within each layer should participate in unlearning. 
LLMs are composed of hierarchical modules, primarily multi-layer perceptrons (MLPs) and multi-head attention (MHA) layers. 
MLPs are central to storing factual knowledge~\cite{nips/MengBAB22}: at layer~$\ell$, the input $\mathbf{x}^\ell$ is transformed as
$
\mathbf{M}^\ell = f(W_K^\ell \mathbf{x}^\ell) W_V^\ell = \mathbf{m}^\ell W_V^\ell,
$
where $\mathbf{M}^\ell$ denotes the layer memory, $W_V^\ell$ is the knowledge matrix, and $f(\cdot)$ generates the intermediate coefficients.  
MHA layers complement MLPs by integrating contextual information across token positions~\cite{emnlp/GevaBFG23}:
\[
\text{MHA}(X) = [\text{Att}_1 \, \| \, \dots \, \| \, \text{Att}_h] W^O,
\]
where $\text{Att}_i$ is the $i$-th head output, $\|$ denotes concatenation, and $W^O$ is the output projection.  
Empirical studies suggest that the MLP and MHA layers concentrate most of the stored knowledge of a model~\cite{nips/MengBAB22}. 
Therefore, we constrain unlearning to these two components.

\paragraph{Summary.}
The three components of \fit work in concert: Redundancy Filtering prevents curvature amplification from repeated similar requests; Importance-Guided Scaling adapts update strengths to suppress directional drift; and Layer Attribution enables stable, targeted forgetting by focusing updates on request-relevant layers.
Together, they form a robust theoretical framework for continual unlearning, balancing forgetting effectiveness, utility preservation, and update stability.

\section{Experimental Configuration Details}\label{Experimental configuration details}
\subsection{Fine-tuned and Retain Models}
\label{Fine-tuned and Retain models}
\begin{figure}[!t]
  \centering
  \subcaptionbox{Retain model\label{subfig:retrain}}[
    0.45\linewidth
  ]{
    \includegraphics[width=\linewidth]{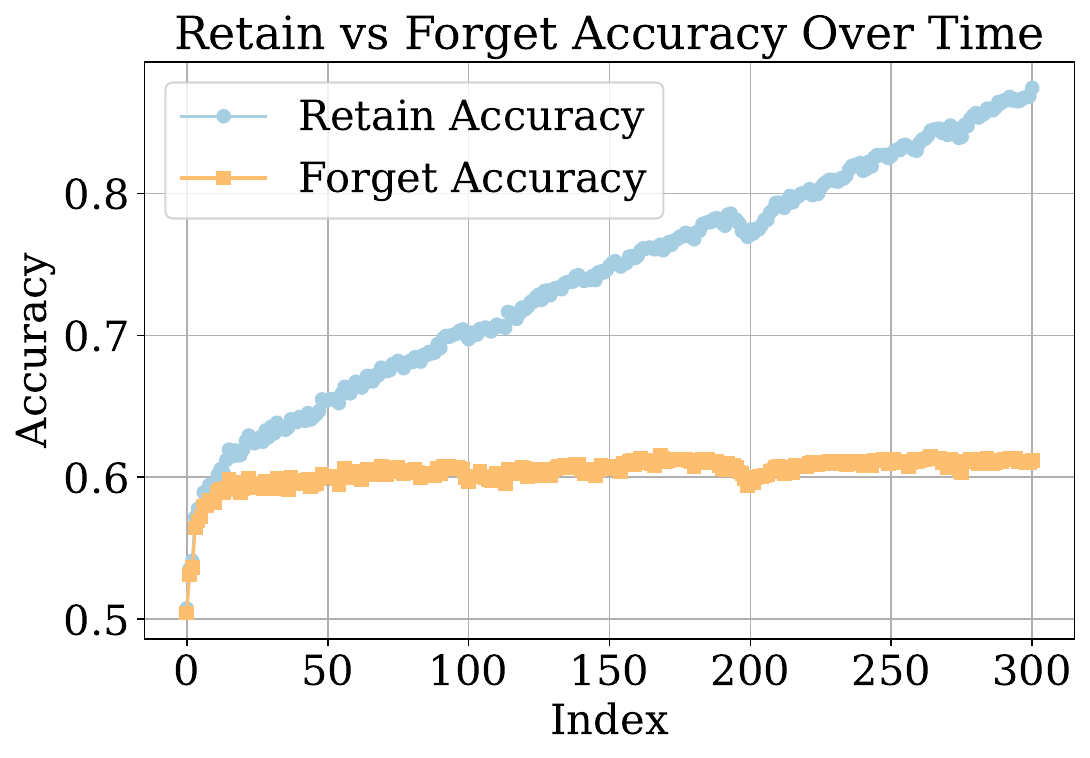}
  }
  \hfill
  \subcaptionbox{Fine-tuned model\label{subfig:finetune}}[
    0.45\linewidth %
  ]{
    \includegraphics[width=\linewidth]{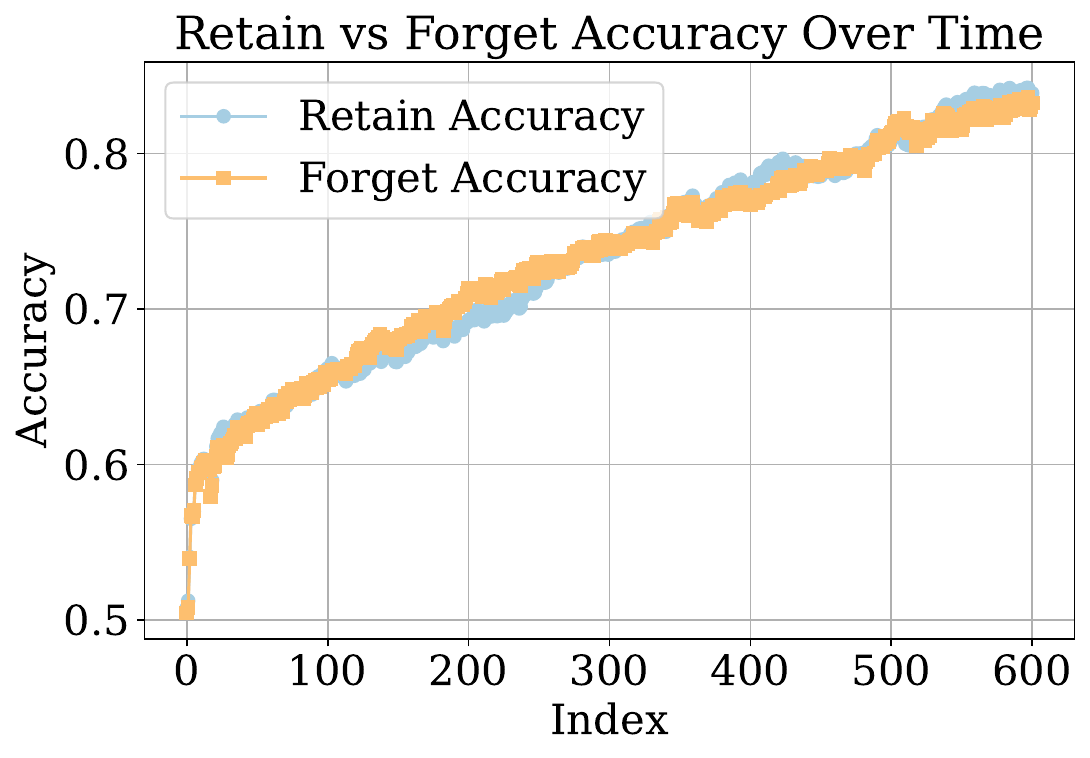}
  }
  \caption{Retain / Forget accuracy under two training regimes}
  \label{fig:retain-forget-llama2}
\end{figure}

Since the pre-training corpus $\mathsf{D}$ is unavailable, retraining directly on the retain subset $\mathsf{D}_r$ is infeasible. 
Following the strategy of~\cite{colm/Maini24}, we construct synthetic counterparts $\mathsf{D}_f$ and $\mathsf{D}_r$, from which we derive a \emph{fine-tuned model} and a \emph{retain model} as practical proxies for the original and retrained models, respectively.
Figure~\ref{fig:retain-forget-llama2} shows that both models exhibit low initial accuracy on their corresponding sets, confirming that these examples were not memorized prior to fine-tuning. Fine-tuning on the full benchmark raises accuracy on both sets. In contrast, tuning only on the retain set progressively widens and then fixes an accuracy gap between retain and forget samples, indicating that the two sets, while carefully curated to be similar, are not identical. In continual unlearning, retain models must be produced sequentially in request order, making it impractical to store a checkpoint after every unlearning step. 
Fortunately, the resulting accuracy curves are smooth and nearly monotonic, so we approximate intermediate baselines by linearly interpolating between the initial fine-tuned model and the first retain checkpoint. For fairness and consistency, at each evaluation checkpoint, all unlearning methods are compared against the same corresponding retain model. This interpolation is an approximation and may introduce small absolute deviations, but it is shared by all compared methods and therefore preserves the fairness of relative comparisons.

\subsection{Experimental Configuration}\label{Experimental Configuration}
\begin{wraptable}{r}{0.40\textwidth}
    \centering
    \vspace{-10pt}
    \caption{Continual unlearning schedule}
    \label{tab:pch_schedule}
    \resizebox{0.36\textwidth}{!}{
    \begin{tabular}{c|cc|c}
    \hline
    \textbf{Requests} ($t$) 
    & $\boldsymbol{\mathsf{D}_f^{(1:t)}}$ 
    & $\boldsymbol{\mathsf{D}_r^{(1:t)}}$ 
    & $\boldsymbol{|\mathsf{D}|}$ \\
    \hline
    0   & 0   & 600 & 600 \\
    60  & 60  & 540 & 600 \\
    120 & 120 & 480 & 600 \\
    180 & 180 & 420 & 600 \\
    240 & 240 & 360 & 600 \\
    300 & 300 & 300 & 600 \\
    \hline
    \end{tabular}
    }
    \vspace{-18pt}
\end{wraptable}
All experiments use consistent settings across datasets, adopting optimizer configurations from~\cite{corr/Touvro23}. We unlearn LLMs with AdamW~\cite{iclr/LoshchilovH19}, using a learning rate of $3.0 \times 10^{-6}$, $\beta_1 = 0.9$, $\beta_2 = 0.95$, and $\epsilon = 10^{-8}$. A cosine learning rate schedule is employed, including a $10\%$ warmup phase and decaying to $10\%$ of the peak rate. Weight decay is set to $0.1$. Our method is conducted on a single NVIDIA H100 GPU, whereas some memory-intensive baselines (\eg, ALKN) use two GPUs.

We evaluate continual unlearning on \textbf{PCH}, where the full dataset is denoted as $\mathsf{D}$ with $|\mathsf{D}|=600$.
Unlearning proceeds sequentially, with one request arriving at a time.
After processing the first $t$ requests, the cumulative forget set is $\mathsf{D}_f^{(1:t)}=\bigcup_{i=1}^{t}\mathcal{D}_f^{(i)}$, and the corresponding retain set is $\mathsf{D}_r^{(1:t)}=\mathsf{D}\setminus \mathsf{D}_f^{(1:t)}$.
We then evaluate forgetting on $\mathsf{D}_f^{(1:t)}$ and utility on $\mathsf{D}_r^{(1:t)}$.
For concise reporting, we record results after every 60 requests (\ie, at $t\in\{0,60,120,180,240,300\}$).
Table~\ref{tab:pch_schedule} summarizes the corresponding set sizes.
All results are reported as the mean over five runs.

\section{Related Work}\label{Related}

\paragraph{Foundations of Machine Unlearning.}  
Machine unlearning has become a critical research direction for addressing privacy, safety, and bias concerns~\cite{acl/YaoCDNWCY24,acl/JangYYCLLS23,icml/PawelczykNL24,icml/LiPGYBGLDGMHLJL24,nips/Liu24,iclr/gao25,iclr/shi24,emnlp/xu25,iclr/ZhangWLWTL00W25,iclr/YuanPDC0L25,icml/xu2025,icml/wuerkaixi2025adaptive,sp/BourtouleCCJTZL21,sp/HuWDX24,sp/Xia25,usenix/long25,usenix/Ye25,ccs/Liu25}.  
Unlearning can be either \emph{exact} or \emph{approximate}~\cite{sp/BourtouleCCJTZL21}. Exact unlearning requires that the resulting model be indistinguishable from one retrained from scratch on the retain set, with all statistical traces of the forget set removed. Approximate unlearning relaxes this requirement to distributional or behavioral similarity, demanding only comparable outputs (\eg, perplexity or accuracy) between unlearned and retain models~\cite{colm/Maini24,iclr/shi24}. For modern LLMs, however, exact unlearning is largely infeasible, as full retraining or partition-based schemes such as SISA~\cite{sp/BourtouleCCJTZL21} are prohibitively expensive. Consequently, approximate unlearning has become the practical choice.  
Yet in the context of LLMs, even state-of-the-art unlearning methods remain vulnerable to adversarial threats such as \emph{malicious unlearning} (attackers submit repetitive deletion requests to degrade model utility~\cite{corr/Barez25,sp/Xia25}), \emph{relearning via fine-tuning}~\cite{acl/LoBC24}, and \emph{quantization attacks} (recovering residual information from low-bit compressed weights~\cite{iclr/ZhangWLWTL00W25}) under continual unlearning setting.

\paragraph{Single-Shot Unlearning.}  
A variety of efficient unlearning strategies has been proposed for LLMs. Gradient ascent and descent methods, including GA and GA+GD, enforce forgetting but may cause utility loss~\cite{acl/YaoCDNWCY24}. Prompt-based methods steer outputs away from sensitive content without parameter updates, reducing computation but often resulting in incomplete forgetting and memory reactivation~\cite{nips/Liu24}. Model editing approaches, such as task arithmetic~\cite{iclr/IlharcoRWSHF23}, AlphaEdit~\cite{corr/Li2505}, and PISCES~\cite{emnlp/gur2025}, which explicitly locate the model regions responsible for forgotten information and are lightweight, targeted, and potentially more robust. However, their effectiveness under realistic, sequential unlearning requests remains largely underexplored.

\paragraph{Continual Unlearning.}  
While single-shot approaches can be effective for isolated deletion events, extending them to continual settings where unlearning requests arrive sequentially often results in catastrophic forgetting and even model collapse. Each new request operates on an already modified model, compounding utility loss and creating unstable dynamics~\cite{corr/Barez25,iclr/shi24}. Recent efforts have explored orthogonal unlearning with LoRA~\cite{iclr/HuSWALWWC22} and out-of-distribution (OOD) detectors to alleviate these issues, but evaluations are typically restricted to a small number of requests on homogeneous datasets such as ScienceQA~\cite{nips/lu22learn} or TOFU~\cite{colm/Maini24}. In more realistic scenarios where the forget and retain sets overlap, OOD detectors suffer sharp accuracy drops, while the LoRA structure may lead to stronger reactivation of forgotten knowledge. ALKN~\cite{icml/wuerkaixi2025adaptive} explores a different direction by providing a theoretical framework for continual unlearning and mitigating accumulative decline and cascading degradation through parameter-level interventions and adaptive modules. CURaTE~\cite{acl/Bae26} takes yet another perspective by formulating continual unlearning as a real-time retrieval-and-refusal process that detects queries related to prior forget requests and selectively abstains, thereby enabling immediate unlearning while preserving the base model parameters and retained knowledge.

\paragraph{Benchmarks.}  
Most unlearning datasets rely on a mix of GPT-generated content and human annotation. TOFU~\cite{colm/Maini24} is fully synthetic, enabling retraining-based baselines. MUSE~\cite{iclr/shi24} leverages authentic corpora such as BBC news and the Harry Potter series, partitioned into forget, retain, and holdout sets. WMDP~\cite{icml/LiPGYBGLDGMHLJL24} targets hazardous capability unlearning with 3668 expert-written multiple-choice questions. RWKU~\cite{nips/JinCWHYL00024} expands adversarial evaluation by combining GPT-4 generation with human review. Despite these advances, existing benchmarks still cover only narrow deletion scenarios such as personal information, copyright, or harmful content, as summarized in Table~\ref{tab:dataset-overview}. 

\begin{figure*}[!htbp]
    \centering
    \subfloat[Fixed Accuracy]{%
        \includegraphics[width=0.24\textwidth]{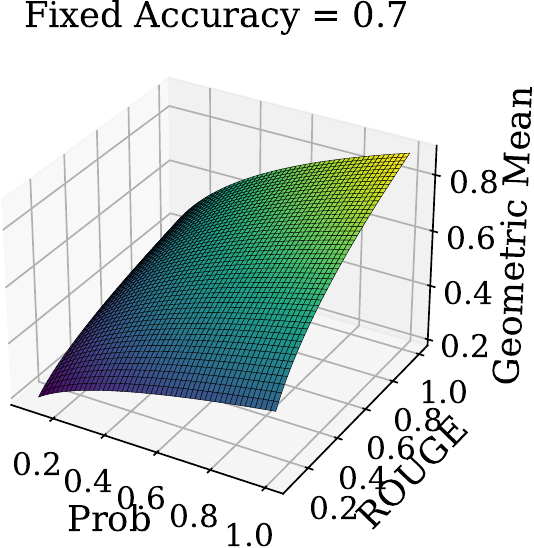}
    }
    \hfill
    \subfloat[Fixed Probability]{%
        \includegraphics[width=0.24\textwidth]{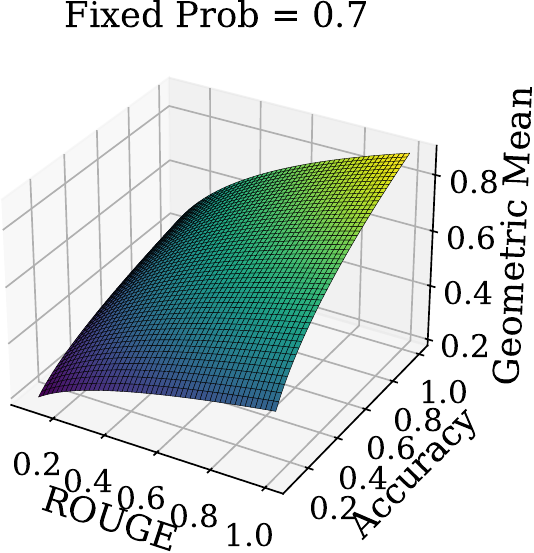}
    }
    \hfill
    \subfloat[Fixed ROUGE-L]{%
        \includegraphics[width=0.24\textwidth]{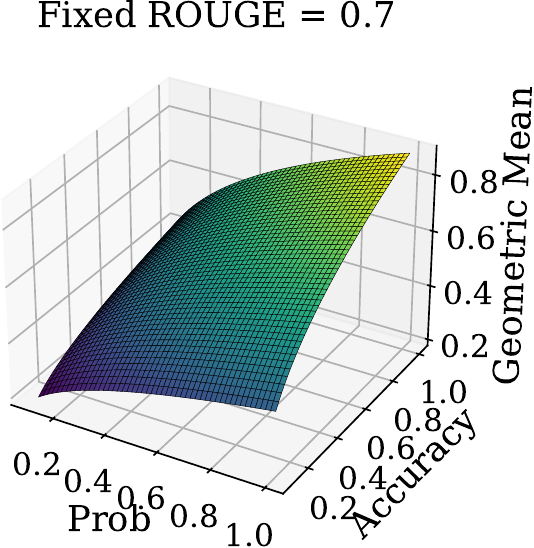}
    }
    \hfill
    \subfloat[F.D. vs. Ratio Curve]{%
        \includegraphics[width=0.24\textwidth, height=3.5cm]{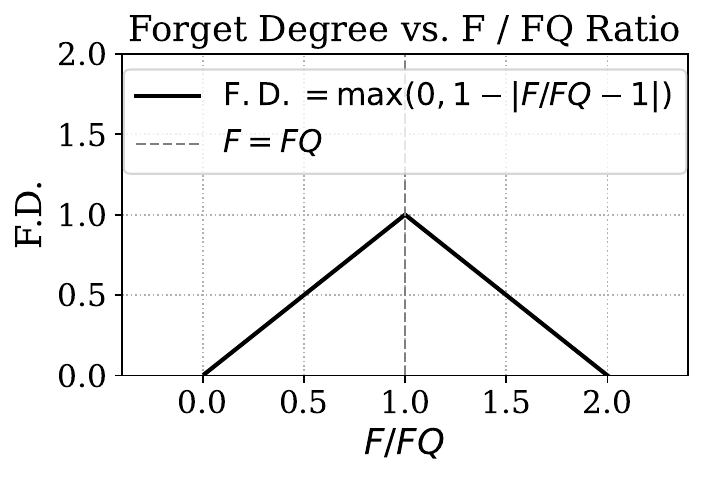}
    }
    \caption{
    \textbf{Geometric‐mean behavior and Forget Degree (F.D.) sensitivity}: 
    Panels (a–c) plot the geometric mean while fixing one component (accuracy/probability/ROUGE-L) at $0.7$, showing that any single ``weak'' component induces a steep decline, confirming the geometric mean as a balanced aggregator. 
    Panel (d) charts F.D.\ as a function of the ratio \(F/FQ\), revealing a symmetric, linear decline from the optimum and demonstrating that F.D.\ is both scale-invariant and easily interpretable. 
    }
    \label{fig:fd-geo-surface}
\end{figure*}

\section{Details of Base Metrics}
\label{app:base-metrics}

This section presents the base metrics in evaluation, including \emph{Probability}, \emph{ROUGE-L}, and \emph{Accuracy}.

\smallskip
\noindent\textbf{Probability.} 
Given a question–answer pair $(q, a)$, 
\[
\operatorname{Prob}(a \mid q) = P(a \mid q)^{1 / |a|}
\]
measures the model's average per-token likelihood, normalized by answer length $|a|$. 
It captures shifts in confidence introduced by unlearning~\cite{colm/Maini24}.

\smallskip
\noindent\textbf{ROUGE-L.} 
This metric quantifies the overlap between the predicted answer $\hat{a}$ and the reference $a$ via the F1 score computed from the length of their longest common subsequence. 
It jointly reflects precision and recall, capturing both exact content preservation and sequence-level similarity.

\smallskip
\noindent\textbf{Accuracy.}
We compute token-level next-token accuracy for each sample under teacher forcing.
For a tokenized sample $\mathbf{x}=(x_1,\ldots,x_T)$, we align predictions with labels by a one-token shift and score positions $t=2,\ldots,T$.
Let $m_t$ be the attention mask from the tokenizer, where $m_t=1$ for non-padding tokens and $m_t=0$ for padding tokens.
The per-sample accuracy is
\[
\operatorname{Acc}(\mathbf{x})=
\frac{1}{\sum_{t=2}^{T} m_t}
\sum_{t=2}^{T} \mathbf{1}[\hat{x}_t = x_t]\cdot m_t.
\]
Here, $T$ is the (padded) sequence length of a \emph{single} sample; we report the dataset-level accuracy by averaging $\operatorname{Acc}(\mathbf{x})$ over all samples.

\section{Analysis of PCH and QA Pairs}{\label{Dataset}}
\begin{table}[!t]
\centering
\caption{Prompts for personal information in \textbf{PCH}}
\label{tab:prompt_personal}
\begin{tcolorbox}[
  colback=blue!2,
  colframe=blue!45!black,
  title=\textbf{Prompt in Personal Information Generation},
  fonttitle=\small\bfseries
]
\textbf{Personal Information.} \\
\textit{Please generate a completely fictional dataset consisting of 200 records. Each record should include the following fields: Name, Age, Address, Occupation, and Description. \\Requirements: (1) All field contents must be purely randomly generated and entirely fictional, with no relation to any real-world data. (2) The Name should consist of commonly used names but be entirely fabricated; Age should be within a realistic range (\eg, 18--80); Address should be constructed using non-existent street names and cities; Occupation and Description must be logically coherent and internally consistent. (3) The data generation process must strictly follow both randomization and rule-based principles to ensure that no pre-trained data or real-world specifics are used.}
\end{tcolorbox}
\end{table}

\subsection{Analysis of PCH}\label{Analysis of PCH}

\begin{wrapfigure}{r}{0.48\textwidth}
    \centering
    \vspace{-10pt}
    \includegraphics[width=0.43\textwidth]{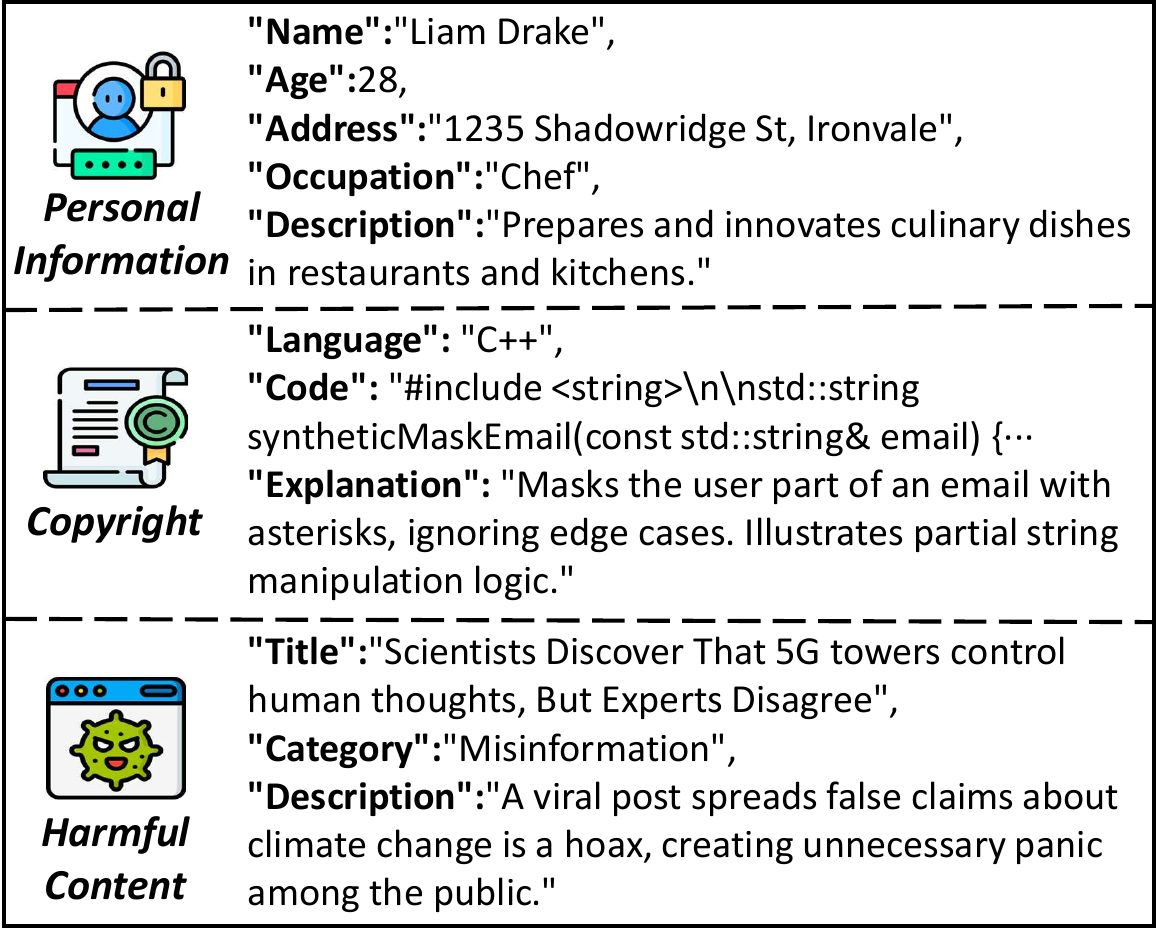}
    \caption{An example from \textbf{PCH} dataset}
    \label{fig:PCH_datatset}
    \vspace{-10pt}
\end{wrapfigure}

\textbf{Personal information}: synthetic individual profiles with attributes such as name, age, and address. As shown in Table~\ref{tab:prompt_personal}, we use structured prompts to synthesize the personal information subset.

\noindent \textbf{Copyright}: machine-generated research papers and code snippets, post-processed to resemble realistic copyrighted material while respecting GPT-4o safeguards. 

\noindent \textbf{Harmful content}: ethically sensitive but permissible text, like misinformation, hate speech, biased statements, conspiracy theories, and manipulative narratives.
We conduct an analysis to verify that the forget and retain sets are distributionally similar yet non-identical, a property that is crucial for evaluating practical unlearning scenarios. All harmful-content samples are synthetic, non-operational, and filtered to avoid actionable instructions or directly harmful procedural content.

\paragraph{Document Length Distribution:}

\begin{figure}[!t]
  \centering
  \begin{subfigure}[t]{0.48\linewidth}
    \centering
    \includegraphics[width=\linewidth]{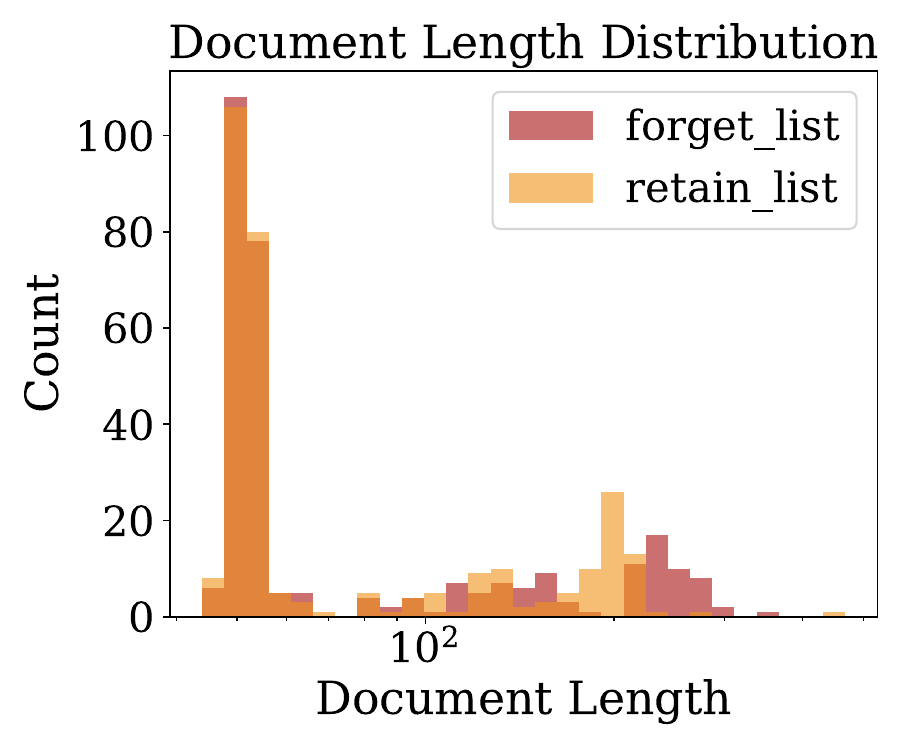}
    \caption{Token-level document length distribution for the forget and retain datasets (log-scaled x-axis)}
    \label{fig:doc-length-dist}
  \end{subfigure}
  \hfill
  \begin{subfigure}[t]{0.48\linewidth}
    \centering
    \includegraphics[width=\linewidth]{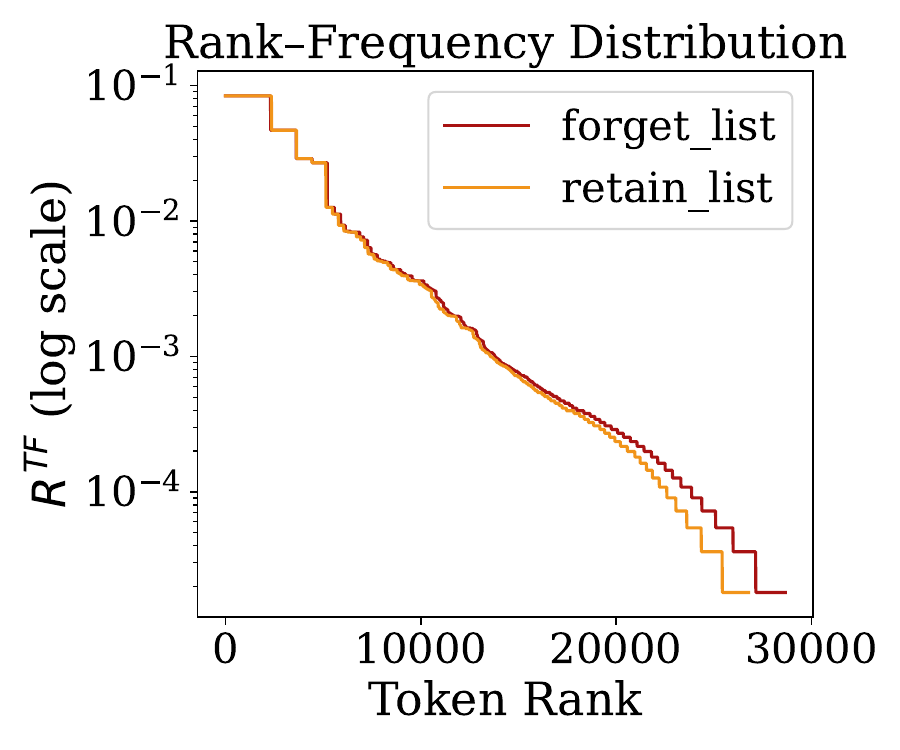}
    \caption{Rank-frequency distribution of tokens from the forget and retain datasets}
    \label{fig:token-rank-frequency}
  \end{subfigure}
  \caption{Distributional comparison of the forget and retain datasets in terms of document length and token frequency}
  \label{fig:dataset-distributions}
\end{figure}

Figure~\ref{fig:doc-length-dist} plots the histogram of document lengths on a logarithmic $x$-axis.  
For a document $d$ with token sequence $t(d)$, its length is
$$
  \ell(d) = \lvert t(d) \rvert .
$$
The empirical length distribution of a collection $\mathcal{C}\in \{forget\ set,retain\ set\}$ is
$$
  P_{\mathcal{C}}(\ell) =
  \frac{1}{\lvert \mathcal{C}\rvert}
  \sum_{d\in\mathcal{C}}
  \mathbf{1}\!\bigl[\ell(d)=\ell\bigr].
$$
Both sets peak at short lengths (\(<100\) tokens).  
The retain set is more concentrated in this region, while the forget set shows a heavier right tail with a sizable share of documents exceeding 200 tokens.  
\paragraph{Token Rank–Frequency Distribution:}
Figure~\ref{fig:token-rank-frequency} presents the rank–frequency distribution of tokens in the forget and retain sets. For each token $\nu$, its frequency is defined:
$$
R^{TF}(\nu) = \frac{\text{count}(\nu)}{\text{total tokens}} .
$$
Tokens are sorted in descending $R^{TF}$, and the curve plots the mapping $r \mapsto R^{TF}(\nu_r)$ on a log scale. The two curves largely overlap across mid- and low-frequency regions, indicating that the forget and retain sets share broadly similar vocabularies. Minor deviations appear only at the extreme tails, reflecting differences in very rare tokens.

\subsection{Analysis of QA Pair}\label{Analysis of QA Pair}

\paragraph{QA Pair Construction}\label{sec:qa-pair-construction}
To enable fine-grained and realistic assessment of unlearning effectiveness, we augment each data sample in the \textbf{PCH} benchmark with a synthetic question--answer (QA) pair. Each question is designed to probe a factual or semantic property unique to the associated sample, while the answer contains specific content that may be the direct target of unlearning requests. We construct QA pairs using a structured prompting template (Table~\ref{tab:prompt_qa}) and ensure they span a wide range of domains, including scientific publications, code snippets, and personal information, to comprehensively assess knowledge memorization. Illustrative examples are provided in Table~\ref{tab:qa-pairs}.

For instance, questions referencing synthetic research articles focus on key findings or arguments unique to the generated text. Code-related QA pairs target both the intent and behavior of program fragments. Personal information questions directly query sensitive details, such as addresses or names, that would be typical candidates for privacy-driven removal. This setting ensures that we can evaluate unlearning performance across realistic use cases.

The design of the QA pairs serves two primary objectives. First, by tightly coupling questions to sample-specific information, we can reliably assess whether unlearning methods effectively remove knowledge pertaining to the forget set without broadly degrading model capabilities. Second, the diversity in question type and domain simulates a practical environment in which deletion requests may span scientific, technical, and personal content.

\begin{table}[!t]
\centering
\caption{Prompt templates for constructing QA pairs}
\label{tab:prompt_qa}
\begin{tcolorbox}[
  colback=teal!2,
  colframe=teal!45!black,
  title=\textbf{Prompt for QA Pair Construction},
  fonttitle=\small\bfseries
]
\textit{You will be provided with a sample. Your goal is to create a question--answer pair that assesses reading comprehension and memorization, ensuring that the question can only be answered using details from the sample. Return a QA Pair with:\\ (1) ``question'': a single specific question that admits only one correct answer and is not answerable from common knowledge; a short span from the sample should suffice to answer it. (2) ``answer'': the exact answer copied verbatim, character-by-character from the sample. Extract the minimal span that fully answers the question.}
\end{tcolorbox}
\end{table}

\begin{table}[!t]
  \caption{Illustrative QA pairs in our evaluation}
  \label{tab:qa-pairs}
  \centering
  \renewcommand{\arraystretch}{1.25}  
  \setlength{\tabcolsep}{6pt}         
  \begin{tabular}{>{\raggedright\arraybackslash}p{0.46\linewidth}%
                  >{\raggedright\arraybackslash}p{0.46\linewidth}}
    \toprule
    \textbf{Question} & \textbf{Answer} \\ \midrule
    What key finding or argument does the paper \emph{``Hypothetical Astronomy Paper: Planet-Scale Magnetic Field Oscillations''} emphasize? &
    Synthetic models indicate that Celora-9’s core, composed of a fictional metal named \emph{Isadrum}, generates oscillatory magnetic pulses. \\ \midrule
    What does the first line of this Java code \texttt{public class Synthetic \ EquationSolver \{} intend to demonstrate? &
    It solves a quadratic equation but only returns one root, ignoring negative or complex possibilities—an example of partial problem solving. \\ \midrule
    What street does Liam Hawthorne live on? &
    7928 Everglow~St. \\ \bottomrule
  \end{tabular}
\end{table}

\section{Additional Result and Efficiency Analysis}{\label{Additional Result and Efficiency Analysi}}
\subsection{Additional Result}\label{Additional result}

\paragraph{Yi-6B, Llama-2-7b-chat-hf, and Llama-3-8B-Instruct.}
Table~\ref{tab:main_results_part2} reports additional results on three backbones. Overall, the same trends hold across models: \fit generally achieves the strongest long-horizon F.D.–R.U. trade-off, especially at later request stages, while aggressive baselines such as GA, and RLabel often collapse as requests accumulate. Conservative methods such as NPO+KL, ALKN, and \(O^3\) preserve utility better, but are generally less stable in forgetting or downstream performance. At 300 requests, \fit achieves the best overall results on Yi-6B and Llama-2-7b-chat-hf, and remains highly competitive on Llama-3-8B-Instruct, where \(O^3\) or ALKN can be stronger in a few intermediate checkpoints. These results suggest that \fit generalizes across different model families.
\begin{table}[!t]
\centering
\caption{Continual unlearning performance across models and request stages}
\label{tab:main_results_part2}
\footnotesize
\setlength{\tabcolsep}{2.8pt}
\renewcommand{\arraystretch}{1.08}
\resizebox{\textwidth}{!}{
\begin{tabular}{lccccccccccccc}
\toprule
\rowcolor{headergray}
\textbf{Method}
& \multicolumn{2}{c}{\textbf{60 req}} 
& \multicolumn{2}{c}{\textbf{120 req}} 
& \multicolumn{2}{c}{\textbf{180 req}} 
& \multicolumn{2}{c}{\textbf{240 req}} 
& \multicolumn{5}{c}{\textbf{300 req}} \\
\cmidrule(lr){2-3}
\cmidrule(lr){4-5}
\cmidrule(lr){6-7}
\cmidrule(lr){8-9}
\cmidrule(lr){10-14}
\rowcolor{subheadergray}
& \textbf{F.D.$\uparrow$} & \textbf{R.U.$\uparrow$} 
& \textbf{F.D.$\uparrow$} & \textbf{R.U.$\uparrow$} 
& \textbf{F.D.$\uparrow$} & \textbf{R.U.$\uparrow$} 
& \textbf{F.D.$\uparrow$} & \textbf{R.U.$\uparrow$} 
& \textbf{F.D.$\uparrow$} & \textbf{R.U.$\uparrow$} 
& \textbf{MMLU$\uparrow$} & \textbf{CSQA$\uparrow$} & \textbf{GSM8K$\uparrow$} \\
\midrule

\rowcolor{modelgray}
\multicolumn{14}{c}{\textbf{Llama-2-7b-chat-hf}} \\

\rowcolor{retainlight}
\textbf{Retain Model}
& \scorestd{1.00}{0.00} & \scorestd{1.00}{0.00}
& \scorestd{1.00}{0.00} & \scorestd{1.00}{0.00}
& \scorestd{1.00}{0.00} & \scorestd{1.00}{0.00}
& \scorestd{1.00}{0.00} & \scorestd{1.00}{0.00}
& \scorestd{1.00}{0.00} & \scorestd{1.00}{0.00}
& \scorestd{47.25}{0.00} & \scorestd{58.89}{0.00} & \scorestd{27.29}{0.00} \\

GA
& \scorestd{0.19}{0.05} & \scorestd{0.18}{0.05}
& \scorestd{0.07}{0.03} & \scorestd{0.07}{0.03}
& \scorestd{0.03}{0.02} & \scorestd{0.07}{0.03}
& \scorestd{0.06}{0.03} & \scorestd{0.08}{0.03}
& \scorestd{0.02}{0.01} & \scorestd{0.03}{0.02}
& \scorestd{27.04}{1.42} & \scorestd{19.41}{1.55} & \scorestd{0.00}{0.00} \\

GA+GD
& \scorestd{0.86}{0.04} & \scorestd{0.87}{0.04}
& \scorestd{0.84}{0.04} & \scorestd{0.82}{0.04}
& \scorestd{0.59}{0.06} & \scorestd{0.58}{0.06}
& \scorestd{0.80}{0.04} & \scorestd{0.74}{0.05}
& \scorestd{0.77}{0.05} & \scorestd{0.72}{0.05}
& \scorestd{41.00}{1.03} & \scorestd{50.15}{1.12} & \scorestd{19.99}{1.26} \\

GA+KL
& \scorestd{0.62}{0.06} & \scorestd{0.65}{0.06}
& \scorestd{0.18}{0.05} & \scorestd{0.16}{0.04}
& \scorestd{0.08}{0.03} & \scorestd{0.06}{0.03}
& \scorestd{0.08}{0.03} & \scorestd{0.04}{0.02}
& \scorestd{0.07}{0.03} & \scorestd{0.03}{0.02}
& \scorestd{27.04}{1.45} & \scorestd{20.39}{1.48} & \scorestd{0.00}{0.00} \\

NPO
& \scorestd{0.74}{0.05} & \scorestd{0.76}{0.05}
& \scorestd{0.63}{0.06} & \scorestd{0.60}{0.06}
& \scorestd{0.42}{0.07} & \scorestd{0.34}{0.07}
& \scorestd{0.59}{0.06} & \scorestd{0.44}{0.07}
& \scorestd{0.52}{0.07} & \scorestd{0.46}{0.07}
& \scorestd{36.56}{1.18} & \scorestd{49.99}{1.16} & \scorestd{5.84}{1.34} \\

NPO+KL
& \scorestd{0.84}{0.04} & \scorestd{0.86}{0.04}
& \scorestd{0.89}{0.03} & \scorestd{0.83}{0.04}
& \scorestd{0.88}{0.03} & \scorestd{0.89}{0.03}
& \scorestd{0.82}{0.04} & \scorestd{0.86}{0.04}
& \scorestd{0.89}{0.03} & \scorestd{0.82}{0.04}
& \scorestd{42.69}{0.94} & \scorestd{51.03}{1.05} & \scorestd{21.20}{1.18} \\

RLabel
& \scorestd{0.32}{0.06} & \scorestd{0.30}{0.06}
& \scorestd{0.22}{0.05} & \scorestd{0.19}{0.05}
& \scorestd{0.15}{0.04} & \scorestd{0.13}{0.04}
& \scorestd{0.11}{0.04} & \scorestd{0.08}{0.03}
& \scorestd{0.09}{0.03} & \scorestd{0.07}{0.03}
& \scorestd{23.12}{1.58} & \scorestd{19.57}{1.56} & \scorestd{0.00}{0.00} \\

PISCES
& \scorestd{0.21}{0.05} & \scorestd{0.22}{0.05}
& \scorestd{0.15}{0.04} & \scorestd{0.14}{0.04}
& \scorestd{0.15}{0.04} & \scorestd{0.16}{0.04}
& \scorestd{0.17}{0.04} & \scorestd{0.12}{0.04}
& \scorestd{0.15}{0.04} & \scorestd{0.14}{0.04}
& \scorestd{24.58}{1.52} & \scorestd{21.58}{1.44} & \scorestd{0.00}{0.00} \\

ALKN
& \scorestd{0.82}{0.04} & \scorestd{0.79}{0.05}
& \scorestd{0.78}{0.05} & \scorestd{0.76}{0.05}
& \scorestd{0.74}{0.05} & \scorestd{0.74}{0.05}
& \scorestd{0.75}{0.05} & \scorestd{0.72}{0.05}
& \scorestd{0.76}{0.05} & \scorestd{0.73}{0.05}
& \scorestd{45.28}{0.72} & \bestscorestd{58.05}{0.54} & \scorestd{24.81}{0.91} \\

$O^3$
& \scorestd{0.87}{0.04} & \scorestd{0.92}{0.03}
& \scorestd{0.84}{0.04} & \scorestd{0.85}{0.04}
& \scorestd{0.81}{0.04} & \scorestd{0.87}{0.04}
& \scorestd{0.80}{0.04} & \scorestd{0.86}{0.04}
& \scorestd{0.64}{0.06} & \scorestd{0.78}{0.05}
& -- & -- & -- \\

\rowcolor{ourslight}
\textbf{Ours}
& \bestscorestd{0.92}{0.03} & \bestscorestd{0.94}{0.02}
& \bestscorestd{0.96}{0.02} & \bestscorestd{0.98}{0.01}
& \bestscorestd{0.95}{0.02} & \bestscorestd{0.98}{0.01}
& \bestscorestd{0.93}{0.03} & \bestscorestd{0.99}{0.01}
& \bestscorestd{0.94}{0.02} & \bestscorestd{0.98}{0.01}
& \bestscorestd{47.03}{0.18} & \scorestd{57.83}{0.62} & \bestscorestd{26.38}{0.58} \\

\midrule

\rowcolor{modelgray}
\multicolumn{14}{c}{\textbf{Llama-3-8B-Instruct}} \\

\rowcolor{retainlight}
\textbf{Retain Model}
& \scorestd{1.00}{0.00} & \scorestd{1.00}{0.00}
& \scorestd{1.00}{0.00} & \scorestd{1.00}{0.00}
& \scorestd{1.00}{0.00} & \scorestd{1.00}{0.00}
& \scorestd{1.00}{0.00} & \scorestd{1.00}{0.00}
& \scorestd{1.00}{0.00} & \scorestd{1.00}{0.00}
& \scorestd{67.04}{0.00} & \scorestd{75.27}{0.00} & \scorestd{76.35}{0.00} \\

GA
& \scorestd{0.00}{0.00} & \scorestd{0.00}{0.00}
& \scorestd{0.00}{0.00} & \scorestd{0.00}{0.00}
& \scorestd{0.00}{0.00} & \scorestd{0.00}{0.00}
& \scorestd{0.00}{0.00} & \scorestd{0.00}{0.00}
& \scorestd{0.00}{0.00} & \scorestd{0.00}{0.00}
& \scorestd{22.95}{1.58} & \scorestd{19.57}{1.56} & \scorestd{0.00}{0.00} \\

GA+GD
& \scorestd{0.02}{0.02} & \scorestd{0.03}{0.02}
& \scorestd{0.15}{0.04} & \scorestd{0.15}{0.04}
& \scorestd{0.43}{0.07} & \scorestd{0.43}{0.07}
& \scorestd{0.08}{0.03} & \scorestd{0.06}{0.03}
& \scorestd{0.24}{0.05} & \scorestd{0.21}{0.05}
& \scorestd{24.57}{1.48} & \scorestd{68.74}{0.98} & \scorestd{1.14}{0.84} \\

GA+KL
& \scorestd{0.00}{0.00} & \scorestd{0.00}{0.00}
& \scorestd{0.00}{0.00} & \scorestd{0.00}{0.00}
& \scorestd{0.00}{0.00} & \scorestd{0.00}{0.00}
& \scorestd{0.00}{0.00} & \scorestd{0.00}{0.00}
& \scorestd{0.00}{0.00} & \scorestd{0.00}{0.00}
& \scorestd{27.13}{1.42} & \scorestd{19.57}{1.54} & \scorestd{0.00}{0.00} \\

NPO
& \scorestd{0.17}{0.04} & \scorestd{0.12}{0.04}
& \scorestd{0.11}{0.04} & \scorestd{0.10}{0.03}
& \scorestd{0.10}{0.03} & \scorestd{0.09}{0.03}
& \scorestd{0.08}{0.03} & \scorestd{0.07}{0.03}
& \scorestd{0.07}{0.03} & \scorestd{0.07}{0.03}
& \scorestd{23.12}{1.55} & \scorestd{19.58}{1.52} & \scorestd{0.00}{0.00} \\

NPO+KL
& \scorestd{0.58}{0.06} & \scorestd{0.68}{0.06}
& \scorestd{0.59}{0.06} & \scorestd{0.69}{0.06}
& \scorestd{0.58}{0.06} & \scorestd{0.66}{0.06}
& \scorestd{0.59}{0.06} & \scorestd{0.65}{0.06}
& \scorestd{0.61}{0.06} & \scorestd{0.65}{0.06}
& \scorestd{65.91}{0.74} & \scorestd{69.56}{0.92} & \scorestd{62.20}{1.02} \\

RLabel
& \scorestd{0.00}{0.00} & \scorestd{0.00}{0.00}
& \scorestd{0.00}{0.00} & \scorestd{0.00}{0.00}
& \scorestd{0.00}{0.00} & \scorestd{0.00}{0.00}
& \scorestd{0.00}{0.00} & \scorestd{0.00}{0.00}
& \scorestd{0.00}{0.00} & \scorestd{0.00}{0.00}
& \scorestd{24.46}{1.50} & \scorestd{19.25}{1.58} & \scorestd{0.00}{0.00} \\

PISCES
& \scorestd{0.00}{0.00} & \scorestd{0.00}{0.00}
& \scorestd{0.00}{0.00} & \scorestd{0.00}{0.00}
& \scorestd{0.00}{0.00} & \scorestd{0.00}{0.00}
& \scorestd{0.00}{0.00} & \scorestd{0.00}{0.00}
& \scorestd{0.00}{0.00} & \scorestd{0.00}{0.00}
& \scorestd{26.12}{1.44} & \scorestd{22.15}{1.42} & \scorestd{0.00}{0.00} \\

ALKN
& \scorestd{0.81}{0.04} & \scorestd{0.77}{0.05}
& \scorestd{0.75}{0.05} & \scorestd{0.76}{0.05}
& \scorestd{0.84}{0.04} & \bestscorestd{0.80}{0.04}
& \scorestd{0.72}{0.05} & \scorestd{0.71}{0.05}
& \scorestd{0.71}{0.05} & \scorestd{0.70}{0.05}
& \scorestd{58.00}{1.06} & \bestscorestd{75.21}{0.04} & \scorestd{60.61}{1.12} \\

$O^3$
& \scorestd{0.89}{0.03} & \bestscorestd{0.88}{0.03}
& \bestscorestd{0.86}{0.04} & \bestscorestd{0.82}{0.04}
& \bestscorestd{0.88}{0.03} & \scorestd{0.78}{0.05}
& \scorestd{0.79}{0.05} & \scorestd{0.72}{0.05}
& \scorestd{0.74}{0.05} & \scorestd{0.70}{0.05}
& -- & -- & -- \\

\rowcolor{ourslight}
\textbf{Ours}
& \bestscorestd{0.92}{0.03} & \scorestd{0.87}{0.04}
& \scorestd{0.85}{0.04} & \bestscorestd{0.82}{0.04}
& \scorestd{0.87}{0.04} & \scorestd{0.79}{0.04}
& \bestscorestd{0.80}{0.04} & \bestscorestd{0.76}{0.04}
& \bestscorestd{0.82}{0.04} & \bestscorestd{0.78}{0.04}
& \bestscorestd{66.45}{0.42} & \scorestd{74.57}{0.54} & \bestscorestd{65.07}{0.86} \\

\midrule

\rowcolor{modelgray}
\multicolumn{14}{c}{\textbf{Yi-6B}} \\

\rowcolor{retainlight}
\textbf{Retain Model}
& \scorestd{1.00}{0.00} & \scorestd{1.00}{0.00}
& \scorestd{1.00}{0.00} & \scorestd{1.00}{0.00}
& \scorestd{1.00}{0.00} & \scorestd{1.00}{0.00}
& \scorestd{1.00}{0.00} & \scorestd{1.00}{0.00}
& \scorestd{1.00}{0.00} & \scorestd{1.00}{0.00}
& \scorestd{64.03}{0.00} & \scorestd{73.55}{0.00} & \scorestd{38.36}{0.00} \\

GA
& \scorestd{0.01}{0.01} & \scorestd{0.02}{0.02}
& \scorestd{0.00}{0.00} & \scorestd{0.00}{0.00}
& \scorestd{0.00}{0.00} & \scorestd{0.00}{0.00}
& \scorestd{0.00}{0.00} & \scorestd{0.00}{0.00}
& \scorestd{0.00}{0.00} & \scorestd{0.00}{0.00}
& \scorestd{27.04}{1.46} & \scorestd{20.56}{1.50} & \scorestd{0.00}{0.00} \\

GA+GD
& \scorestd{0.52}{0.07} & \scorestd{0.62}{0.06}
& \scorestd{0.64}{0.06} & \scorestd{0.75}{0.05}
& \scorestd{0.62}{0.06} & \scorestd{0.69}{0.05}
& \scorestd{0.63}{0.06} & \scorestd{0.63}{0.06}
& \scorestd{0.67}{0.06} & \scorestd{0.64}{0.06}
& \scorestd{45.99}{1.12} & \scorestd{65.56}{1.02} & \scorestd{17.59}{1.28} \\

GA+KL
& \scorestd{0.18}{0.04} & \scorestd{0.28}{0.06}
& \scorestd{0.02}{0.02} & \scorestd{0.01}{0.01}
& \scorestd{0.00}{0.00} & \scorestd{0.00}{0.00}
& \scorestd{0.00}{0.00} & \scorestd{0.00}{0.00}
& \scorestd{0.00}{0.00} & \scorestd{0.00}{0.00}
& \scorestd{24.46}{1.52} & \scorestd{20.88}{1.48} & \scorestd{0.00}{0.00} \\

NPO
& \scorestd{0.57}{0.06} & \scorestd{0.62}{0.06}
& \scorestd{0.43}{0.07} & \scorestd{0.47}{0.07}
& \scorestd{0.18}{0.05} & \scorestd{0.17}{0.04}
& \scorestd{0.11}{0.04} & \scorestd{0.11}{0.04}
& \scorestd{0.04}{0.02} & \scorestd{0.07}{0.03}
& \scorestd{24.35}{1.54} & \scorestd{20.64}{1.50} & \scorestd{0.00}{0.00} \\

NPO+KL
& \scorestd{0.78}{0.05} & \scorestd{0.81}{0.04}
& \scorestd{0.76}{0.05} & \scorestd{0.74}{0.05}
& \scorestd{0.67}{0.06} & \scorestd{0.76}{0.05}
& \scorestd{0.68}{0.06} & \scorestd{0.69}{0.06}
& \scorestd{0.69}{0.06} & \scorestd{0.69}{0.06}
& \scorestd{57.40}{0.96} & \scorestd{66.34}{0.98} & \scorestd{31.91}{1.08} \\

RLabel
& \scorestd{0.16}{0.04} & \scorestd{0.19}{0.04}
& \scorestd{0.02}{0.02} & \scorestd{0.02}{0.02}
& \scorestd{0.01}{0.01} & \scorestd{0.01}{0.01}
& \scorestd{0.01}{0.01} & \scorestd{0.01}{0.01}
& \scorestd{0.01}{0.01} & \scorestd{0.01}{0.01}
& \scorestd{23.12}{1.58} & \scorestd{19.57}{1.56} & \scorestd{0.00}{0.00} \\

PISCES
& \scorestd{0.12}{0.04} & \scorestd{0.08}{0.03}
& \scorestd{0.00}{0.00} & \scorestd{0.00}{0.00}
& \scorestd{0.00}{0.00} & \scorestd{0.00}{0.00}
& \scorestd{0.00}{0.00} & \scorestd{0.00}{0.00}
& \scorestd{0.00}{0.00} & \scorestd{0.00}{0.00}
& \scorestd{25.57}{1.46} & \scorestd{20.88}{1.48} & \scorestd{0.00}{0.00} \\

ALKN
& \scorestd{0.87}{0.04} & \scorestd{0.85}{0.04}
& \scorestd{0.76}{0.05} & \scorestd{0.77}{0.05}
& \scorestd{0.81}{0.04} & \scorestd{0.82}{0.04}
& \scorestd{0.77}{0.05} & \scorestd{0.75}{0.05}
& \scorestd{0.76}{0.05} & \scorestd{0.74}{0.05}
& \scorestd{58.52}{0.92} & \scorestd{66.34}{0.96} & \scorestd{32.14}{1.04} \\

$O^3$
& \scorestd{0.93}{0.03} & \bestscorestd{0.87}{0.03}
& \scorestd{0.90}{0.03} & \scorestd{0.82}{0.04}
& \scorestd{0.89}{0.03} & \scorestd{0.82}{0.04}
& \scorestd{0.85}{0.04} & \bestscorestd{0.83}{0.04}
& \scorestd{0.89}{0.03} & \scorestd{0.80}{0.04}
& -- & -- & -- \\

\rowcolor{ourslight}
\textbf{Ours}
& \bestscorestd{0.95}{0.02} & \scorestd{0.85}{0.04}
& \bestscorestd{0.94}{0.02} & \bestscorestd{0.86}{0.03}
& \bestscorestd{0.94}{0.02} & \bestscorestd{0.85}{0.03}
& \bestscorestd{0.92}{0.03} & \bestscorestd{0.83}{0.04}
& \bestscorestd{0.91}{0.03} & \bestscorestd{0.82}{0.04}
& \bestscorestd{63.52}{0.38} & \bestscorestd{72.90}{0.52} & \bestscorestd{37.58}{0.60} \\

\bottomrule
\end{tabular}
}
\end{table}

\subsection{Efficiency Analysis}
\begin{wraptable}{r}{0.7\textwidth}
    \centering
    \vspace{-10pt}
    \caption{GPU memory usage on Llama-2-7b-chat-hf}
    \label{tab:params-llama2}
    \setlength{\tabcolsep}{2.2pt}
    \renewcommand{\arraystretch}{0.92}
    \resizebox{0.7\textwidth}{!}{
    \begin{tabular}{lcccccccccc}
    \toprule
    \textbf{Method} 
    & GA & GA+GD & GA+KL & NPO & NPO+KL & RLabel & PISCES & $O^3$ & ALKN & \fit \\
    \midrule
    \textbf{GPU Memory}
    & 81.71 & 88.43 & 89.07 & 81.71 & 86.72 & 79.20 & 27.53 & 22.14 & 140.91 & 41.84 \\
    \bottomrule
    \end{tabular}
    }
    \vspace{-10pt}
\end{wraptable}
We report the number of model parameters updated during unlearning as a direct measure of computational efficiency.
Table~\ref{tab:params-llama2} summarizes the results on Llama-2-7b-chat-hf.
LoRA-based methods~\cite{iclr/gao25} achieve efficiency by limiting the update scope but often exclude critical parameters associated with the forget set, resulting in incomplete forgetting and potential knowledge reactivation.
In contrast, our layer-selection strategy updates fewer than one-quarter of the parameters required for full-model retraining, yet achieves stronger forgetting than $O^3$ and higher robustness than LoRA.
By concentrating updates on layers most responsible for the forgotten content, our targeted attribution mechanism attains an effective trade-off between computational efficiency and unlearning stability.

\section{Threat Model}\label{threat}
\begin{wrapfigure}{r}{0.55\textwidth}
    \centering
    \vspace{-10pt}
    \includegraphics[width=0.50\textwidth]{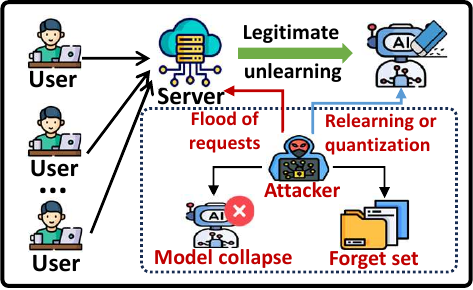}
    \caption{Threat model for continual unlearning: \emph{malicious unlearning} or DoS attacks, \emph{relearning}, and \emph{quantization} attacks}
    \label{fig:attack}
    \vspace{-10pt}
\end{wrapfigure}

\paragraph{Attacker's Goal.}
Figure~\ref{fig:attack} depicts the interaction among the server, legitimate users, and an adversary under the continual unlearning setting.
The server hosts an LLM, answers inference queries, and processes unlearning requests.

The adversary masquerades as a normal user and can mount two types of attacks: 
i) \emph{Malicious unlearning}: 
A rapid stream of unlearning requests forces repeated updates, inducing catastrophic forgetting and degrading performance far beyond the designated forget set (\emph{cf.} a ``denial-of-service'' attack on model utility), or 
ii) \emph{Post-unlearning recovery}:
After the unlearned model is deployed, the attacker attempts to restore the erased knowledge through, \eg, \emph{relearning via fine-tuning}~\cite{acl/LoBC24, corr/Barez25}, or \emph{quantization attacks} that compress the model to low-bit precision, amplifying residual memorization and making it easier to extract sensitive information~\cite{iclr/ZhangWLWTL00W25}.
\paragraph{Attacker's Capabilities.}
Both \emph{malicious unlearning} and \emph{relearning} attacks often assume a \emph{black-box} setting.
Malicious unlearning further assumes that the adversary can issue a burst of unlearning requests in a short time window, inducing catastrophic forgetting and model collapse.

In contrast, relearning assumes that an adversary can obtain auxiliary data and fine-tune the unlearned model. Such data may include (i) a subset of the forget set, or in the worst case the entire forget set~\cite{corr/Barez25,icml/xu2025}, which we do not consider since the forget set is assumed to be private and not observable to the adversary; (ii) the retain set or other samples drawn from a distribution similar to the forget set~\cite{icml/xu2025}; or (iii) completely unrelated, out-of-distribution data~\cite{icml/xu2025}.

\emph{Quantization attacks}~\cite{iclr/ZhangWLWTL00W25} work in a white-box setting with full access to model parameters so that the attacker can subject the weights to aggressive low-bit quantization. 
As the weights of the original model and the unlearned model typically differ only slightly, the quantizer rounds both to the same value, nullifying the effect of unlearning, \eg, $2.301235$ (original) vs. $2.412567$ (unlearned) $\rightarrow$ $2.2858$ after \texttt{int4}, thus restoring much of the unlearned information due to ``weights collision.''

Threats that require altering server infrastructure, modifying the unlearning algorithm, or intercepting private communications between the server and its users are out of scope.

\newpage

\end{document}